\documentclass[journal]{IEEEtran}
\usepackage{multirow}
\usepackage[table,xcdraw]{xcolor}
\usepackage{graphicx}
\usepackage{amsmath}
\usepackage{amsfonts}
\usepackage{algorithm}
\usepackage{amsthm,amssymb}
\usepackage{mathrsfs}
\usepackage{algorithmic}
\usepackage{makecell}%解决公式在表格中位置太小的问题
\usepackage{easyReview}
\usepackage{booktabs}%解决表格横向粗细问题
\setcellgapes{3pt}

\setcellgapes{3pt}

\usepackage[justification=centering]{caption}
\usepackage{color}
\usepackage[subfigure]{tocloft}
\usepackage{subfigure}
\usepackage{tabularx}
\usepackage{amsmath, bm}
\usepackage{hyperref}
\hypersetup{hidelinks,
colorlinks=true,
allcolors=blue,
pdfstartview=Fit,
breaklinks=true}
\usepackage{cite}

\ifCLASSINFOpdf
\else
\fi
\begin{document}
	\title{GAI-Enabled Explainable Personalized Federated Semi-Supervised Learning}
	
	\author{Yubo Peng, \textit{Student Member, IEEE}, Feibo Jiang, \textit{Senior Member, IEEE}, Li Dong, Kezhi Wang, \textit{Senior Member, IEEE}, and Kun Yang, \textit{Fellow, IEEE}
		\thanks{
			{This work was supported in part by the
				National Natural Science Foundation of China under Grant 41904127 and 62132004. in part by the Hunan Provincial Natural Science Foundation of China under Grant 2024JJ5270, in part by the Open Project of Xiangjiang Laboratory under Grant 22XJ03011, and in part by the Scientific Research Fund of Hunan Provincial Education Department under Grant 22B0663.
				
				Yubo Peng (ybpeng@smail.nju.edu.cn) and Kun Yang (kyang@ieee.org) are with the State Key Laboratory of Novel Software Technology, Nanjing University, Nanjing, China, and the School of Intelligent Software and Engineering, Nanjing University (Suzhou Campus), Suzhou, China. 
				
				Feibo Jiang (jiangfb@hunnu.edu.cn) is with the School of Information Science and Engineering, Hunan Normal University, Changsha, China. 

                Li Dong (Dlj2017@hunnu.edu.cn) is with Changsha Social Laboratory of Artificial Intelligence, Hunan University of Technology and Business, Changsha, China.
				
				Kezhi Wang (Kezhi.Wang@brunel.ac.uk) is with the Department of Computer Science, Brunel University London, UK.
			}
			%	\thanks{The paper was submitted on \today.}\\
		}
	}

\markboth{Submitted for Review}%
{Shell \MakeLowercase{\textit{et al.}}: Bare Demo of IEEEtran.cls for IEEE Journals}



\maketitle 

\begin{abstract}
Federated learning (FL) is a commonly distributed algorithm for mobile users (MUs) training artificial intelligence (AI) models, however, several challenges arise when applying FL to real-world scenarios, such as \textit{label scarcity}, \textit{non-IID data}, and \textit{unexplainability}. As a result, we propose  an explainable personalized FL framework, called XPFL.
First, we introduce a generative AI (GAI) assisted personalized federated semi-supervised learning, called GFed. Particularly, in local training, we utilize a GAI model to learn from large unlabeled data and apply knowledge distillation-based semi-supervised learning to train the local FL model using the knowledge acquired from the GAI model. 
In global aggregation, we obtain the new local FL model by fusing the local and global FL models in specific proportions, allowing each local model to incorporate knowledge from others while preserving its personalized characteristics.
Second, we propose an explainable AI mechanism for FL, named XFed. Specifically, in local training, we apply a decision tree to match the input and output of the local FL model. In global aggregation, we utilize t-distributed stochastic neighbor embedding (t-SNE) to visualize the local models before and after aggregation.
Finally, simulation results validate the effectiveness of the proposed XPFL framework.
\end{abstract}


% \fi
\IEEEpeerreviewmaketitle

\section{Introduction}
In the current era of big data and deep learning (DL), with the pervasiveness of Internet of Things technology, the data generated by various mobile users (MUs) has increased dramatically \cite{lakshmanna2022review}. 
Many fields utilize DL to process massive data for efficiency, such as finance, medical care, retail, etc. Since most of these data are sensitive to personal information and business secrets \cite{yang2022privacy}, privacy security has become a critical issue.
However, all data is required to be stored centrally for training in traditional centralized learning (CL), which increases the risk of data leakage. 
As a solution, federated learning (FL) comes into being, which can realize model training and updating under the premise of protecting user privacy and data security \cite{ghimire2022recent}.

The idea of FL is to distribute the model training and update process to MUs, thus avoiding privacy issues in the centralized storage and transmission of data \cite{yang2019federated}.
Specifically, FL enables several MUs and a central server to perform training collaboratively only by sharing model parameters, rather than transmitting large amounts of raw training data. Therefore, FL is viewed as a more efficient and secure machine learning method and can be applied in various application scenarios, such as smart homes, healthcare, and financial services.
%For instance, in \cite{wazzeh2022privacy}, to protect the user privacy, a novel FL based continuous authentication mechanism was introduced for mobile and IoT devices. The authors in \cite{singh2022framework} proposed Blockchain and FL-enabled secure architecture for privacy-preserving in smart healthcare. In \cite{suzumura2022federated}, the authors leveraged FL as well as graph learning approaches to train a DL model that could detect the financial crime risk accurately.

Although FL has shown huge potential and application prospects in terms of data privacy and security, there are still some challenges when FL is applied in real scenarios:
\begin{enumerate}
	\item {\it{Label scarcity}}:
	In practical settings, labeled data on MUs is often scarce \cite{yu2021fedhar}. This scarcity arises from the vast amounts of data generated by interactions with MUs, such as photos, text inputs, and physiological measurements from wearable technology. It is impractical to expect users to label all such data. Moreover, in domains such as finance (e.g., risk management, credit assessment) and healthcare (e.g., disease diagnosis, health monitoring), data often requires expert knowledge for accurate labeling. Unfortunately, most existing FL frameworks, including FedAvg and FedSGD [2], rely on supervised learning and are unable to effectively leverage unlabeled data.
	
	\item {\it{Non-IID data}}:
	FL participants are typically heterogeneous, reflecting variations in local environments and usage patterns. Consequently, the local data across different MUs is often non-independent and identically distributed (non-IID). The size and distribution of local datasets can vary significantly, and no single local dataset may represent the overall distribution. Thus, the assumption that training data is uniformly distributed across MUs does not generally hold in FL \cite{chen2019communication}. This heterogeneity complicates the training process and can hinder the convergence of FL models.
	
	\item {\it{Unexplainability}}:
	DL models, while highly accurate, often function as ``black boxes," making their outputs difficult to interpret \cite{raza2022designing}. In critical fields such as healthcare and finance, where decisions may impact human lives and property, explainability is crucial. Therefore, despite FL has advantages in privacy and security, the lack of explainability remains a significant barrier to its broader application.
\end{enumerate}

Generative artificial intelligence (GAI) \cite{jiang2024large3d} represents a recent advancement in DL technology, with one of its key strengths being the ability to learn effectively from large volumes of unlabeled data. In contrast to traditional discriminative DL models, which depend heavily on labeled data for supervised learning, GAI models can capture the underlying structures and distributions of the data, enabling them to generate new examples that closely resemble the original dataset. This capability makes GAI particularly valuable in scenarios where labeled data is scarce or prohibitively expensive to obtain. As a result, to address the previously mentioned challenges, we propose an explainable personalized FL (XPFL) framework, in which we introduce a GAI-assisted federated semi-supervised learning (GFed) algorithm to solve the issues of label scarcity and non-IID. In addition, we design an explainable artificial intelligence (XAI)--based FL (XFed) to enhance the explainability of both the local FL model and global aggregation. The specific contributions are as follows:
\begin{enumerate}
	\item 
	\emph{GAI-assisted semi-supervised learning:}  In GFed, we utilize semi-supervised learning to train the local FL model during the local training phase. Specifically, we design a GAI-based autoencoder (GAE) to learn from vast amounts of unlabeled data through unsupervised learning. Simultaneously, the local FL model is trained on the limited labeled data using supervised learning. The GAE then transfers the learned knowledge to the local FL model through knowledge distillation (KD)-based semi-supervised learning, thereby addressing the issue of \textit{label scarcity}.
	
	\item 
	\emph{Personalized Global Aggregation:}
	In GFed, we update the local FL model by integrating it with the global model in proportions determined by the difference in weights between the two models. This process enables each local model to acquire knowledge from parameter aggregation while retaining its personalized characteristics, thus mitigating the challenges posed by \textit{non-IID data}.
	
	\item 
	\emph{Explainable Local Model:} In XFed, we employ a decision tree (DT) as a white-box model to approximate the input-output behavior of the local FL model (black-box model). The DT is trained to produce the same output as the local model for identical inputs, serving as an efficient explainer for the predictions made by the local FL model.
	
	\item 
	\emph{Visual Global Aggregation:}
	In XFed, we apply t-distributed Stochastic Neighbor Embedding (t-SNE) \cite{van2008visualizing} to visualize the global aggregation process. By comparing the visualization results before and after model updates, we can directly observe the changes in each local model during global aggregation. Through the combined use of DT and t-SNE,  we can ensure \textit{explainability} across both the local model and global aggregation.
\end{enumerate}

The remainder of this paper is structured as follows. 
Section II outlines the current literature in this field, focusing on FL and its solutions to label scarcity, client heterogeneity, and FL explainability. 
Section III introduces the system model.
Section IV describes the details of the proposed XPFL, including the GFed and XFed methods.
Section V presents the simulation settings and results for evaluating the proposed XPFL framework.
Finally, Section VI concludes this paper.

\section{Related Work}
This section introduces related literature on FL that focuses on solving the three challenges, label scarcity, client heterogeneity, and explainability.

\subsection{FL for Label Scarcity}
Label scarcity is a common challenge in FL settings \cite{ji2021emerging}. To tackle this issue, researchers have increasingly explored semi-supervised and unsupervised learning techniques. For instance, Tsouvalas \textit{et al}. \cite{tsouvalas2022privacy} proposed a privacy-preserving and data-efficient speech emotion recognition approach that combines self-training with FL to leverage both labeled and unlabeled data. Yu \textit{et al}. \cite{yu2021fedhar} introduced an online FL algorithm that computes unsupervised gradients using consistency training for clients lacking labeled data. Similarly, Dong \textit{et al}. \cite{dong2022federated} developed a federated partially supervised learning framework to address the scarcity of labeled data across MUs.

Unlike these approaches, the proposed GAI-assisted semi-supervised learning not only fully exploits the potential knowledge of unlabeled data, but also enhances the performance of the local FL model by integrating a personalized FL scheme.

\subsection{FL for Non-IID data}
Client heterogeneity in FL often manifests through non-IID local data, which can hinder model convergence and degrade performance \cite{zhang2021adaptive}. To address this challenge, researchers have proposed several approaches. Tursunboev \textit{et al}. \cite{tursunboev2022hierarchical} introduced a hierarchical FL algorithm that utilizes edge servers at base stations as intermediate aggregators, employing shared data to mitigate non-IID effects. Wu \textit{et al}. \cite{wu2022node} proposed a probabilistic node selection framework (FedPNS), which dynamically adjusts the probability of selecting each node based on the optimal aggregation output, effectively addressing non-IID challenges. Similarly, Gao \textit{et al}. \cite{gao2022feddc} developed an FL algorithm with local drift decoupling and correction (FedDC), where clients track the difference between local and global model parameters using an auxiliary local drift variable.

The previous approaches have focused on improving the generalization of global models on local data and mitigating the impact of non-IID data, but our proposed personalized FL approach updates local FL model parameters by fusing both local and global FL models. This method allows local FL models to benefit from the global aggregation while maintaining personalization for each client’s unique data distribution.

\subsection{Explainability in FL}
XAI encompasses techniques that help humans understand and interpret the decision-making processes of AI systems, such as DL models. XAI is particularly important in fields like healthcare, manufacturing, and automotive industries. Consequently, researchers have increasingly focused on integrating XAI with FL.
For instance, Huong \textit{et al}. \cite{huong2022federated} proposed the FedeX architecture based on XAI, enabling users to interpret why a model classified an instance as abnormal. Nasiri \textit{et al}. \cite{nasiri2021wearable} applied FL-based methods and case-based reasoning (CBR) to create a wearable XAI framework. Similarly, Pedrycz \textit{et al}. \cite{pedrycz2021design} introduced an FL approach for learning Takagi-Sugeno-Kang fuzzy rule-based systems (TSK-FRBSs), which serve as XAI models for regression tasks, allowing for easy interpretation of the model’s decision-making process.

Unlike the above works that solely focus on FL with XAI to explain the prediction results of local models, our approach simultaneously addresses the explainability of the local models and the global aggregation process.

\section{System Model}
\begin{figure}[htbp]
	\centering
	\includegraphics[width=8.5cm]{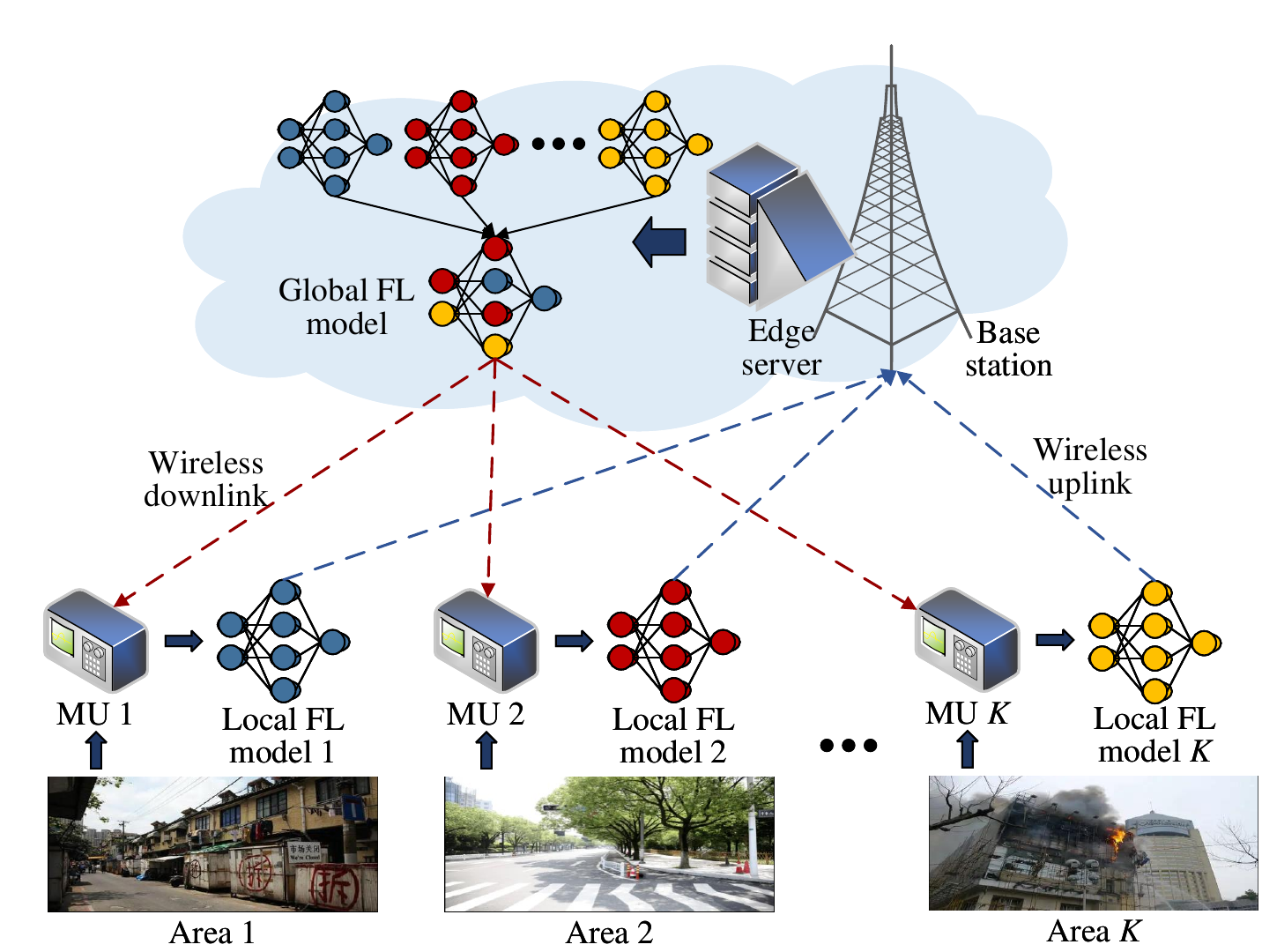}
	\caption{The illustration of the considered system model.}
	\label{fig:system}
\end{figure}

As illustrated in Fig. \ref{fig:system}, we consider a cellular network where a base station (BS) equipped with an edge server and a set $\mathcal{K}$ of $K$ MUs collaboratively employ a FL algorithm for data training and inference. FL allows MUs to train local models using their datasets and update these models by exchanging parameters with the BS. For instance, in a pre-disaster scenario, MUs for monitoring can collect image data from the surveillance area and train a shared DL model using the FL framework.
It is important to note that most of the collected data is unlabeled, with only a small portion being labeled, as the high cost of labeling makes it prohibitive for MUs \cite{yu2021fedhar}. However, in disaster situations where decisions may be life-critical, ensuring the explainability of the FL models is crucial for enhancing the reliability and trustworthiness of model predictions.
\subsection{FL Model}
In the considered FL model, each MU $k$ has a labeled data set, $\mathcal{D}_k^\text{L}=\{(x_{k,i}^\text{L},y_{k,i}^\text{L})\}_{i=1}^{D_k^\text{L}}$, where $x_{k,i}^\text{L}$ is the $i$-th labeled data sample, $y_{k,i}^\text{L}$ is the corresponding label, and $D_k^\text{L}$ is the size of the labeled data set. 
In addition, there is also an unlabeled data set, $\mathcal{D}_k^\text{U}=\{x_{k,i}^\text{U}\}_{i=1}^{D_k^\text{U}}$, where $x_{k,i}^\text{U}$ is the $i$-th unlabeled data sample and $D_k^\text{U}$ is the size of the unlabeled data set. Hence, the local data set of the MU $k$ could be denoted as $\mathcal{D}_k = \mathcal{D}_k^\text{L} \bigcup \mathcal{D}_k^\text{U}$, and the data size is $D_k = D_k^\text{L}+ D_k^\text{U}$.
Note the local data of different MUs may be non-IID, which depends on the monitoring area and the usage pattern of MUs. 

Since the traditional FL usually relies on supervised learning based on the labeled data of MUs, we don't consider the unlabeled data temporarily in this subsection. In the $t$-th communication round of FL, the local loss of the MU $k$ over the local labeled data set $\mathcal{D}_k^\text{L}$ can be calculated as:
\begin{equation}
	F_{k,t}^\text{L}\left( {w}_{k,t} \right) = \frac{1}{D_k^\text{L}}{\sum\limits_{i=1}^{D_k^\text{L}}{f^\text{L}\left( {w}_{k,t}; x_{k,i}^\text{L},y_{k,i}^\text{L}\right)}},
\end{equation}
where ${w}_{k,t}$ represents the local FL model of the MU $k$ in the $t$-th communication round.

After local training, all local FL models $w_{k,t}$ are uploaded to the BS and merged as the global FL model $w_{g,t}$ by parameter aggregation, ensuring data privacy and security. The global aggregation could be described by:
\begin{equation}\label{eq:aggregation}
	w_{g,t}=\frac{1}{D^\text{L}}\sum\limits_{k=1}^{K} D_k^\text{L} w_{k,t},
\end{equation}
where $w_{g,t}$ denotes the global FL model in the $t$-th communication round. Note that $w_{g,t}$ has the same architecture as $w_{n,t}$. $D^\text{L}=\sum\limits_{k=1}^K D_k^\text{L}$ represents the total size of all local data sets.

The traditional FL algorithm aims to obtain a global FL model which could minimize the local loss of each MU $k$ on its labeled data set. The standard FL objective is given as:
\begin{equation}\label{eq:FL_obj1}
	F_{g,t}^\text{L}\left( w_{g,t} \right) = \frac{1}{K}{\sum\limits_{k = 1}^{K}{F_{k,t}^\text{L}\left(w_{g,t}\right)}}.
\end{equation}

Our goal, however, is to achieve personalized FL, which seeks to minimize the local loss for each MU $k$ using the local model $w_{k,t}$, rather than relying solely on the global model $w_{g,t}$. Therefore, the objective for personalized FL can be expressed as:
\begin{equation}\label{eq:FL_obj2}
	F_{g,t}^\text{L}\left( w_{g,t} \right) = \frac{1}{K}{\sum\limits_{k = 1}^{K}{F_{k,t}^\text{L}\left(w_{k,t}\right)}}.
\end{equation}

\section{Methods}
\begin{figure*}[htbp]
	\centering
	\includegraphics[width=18cm]{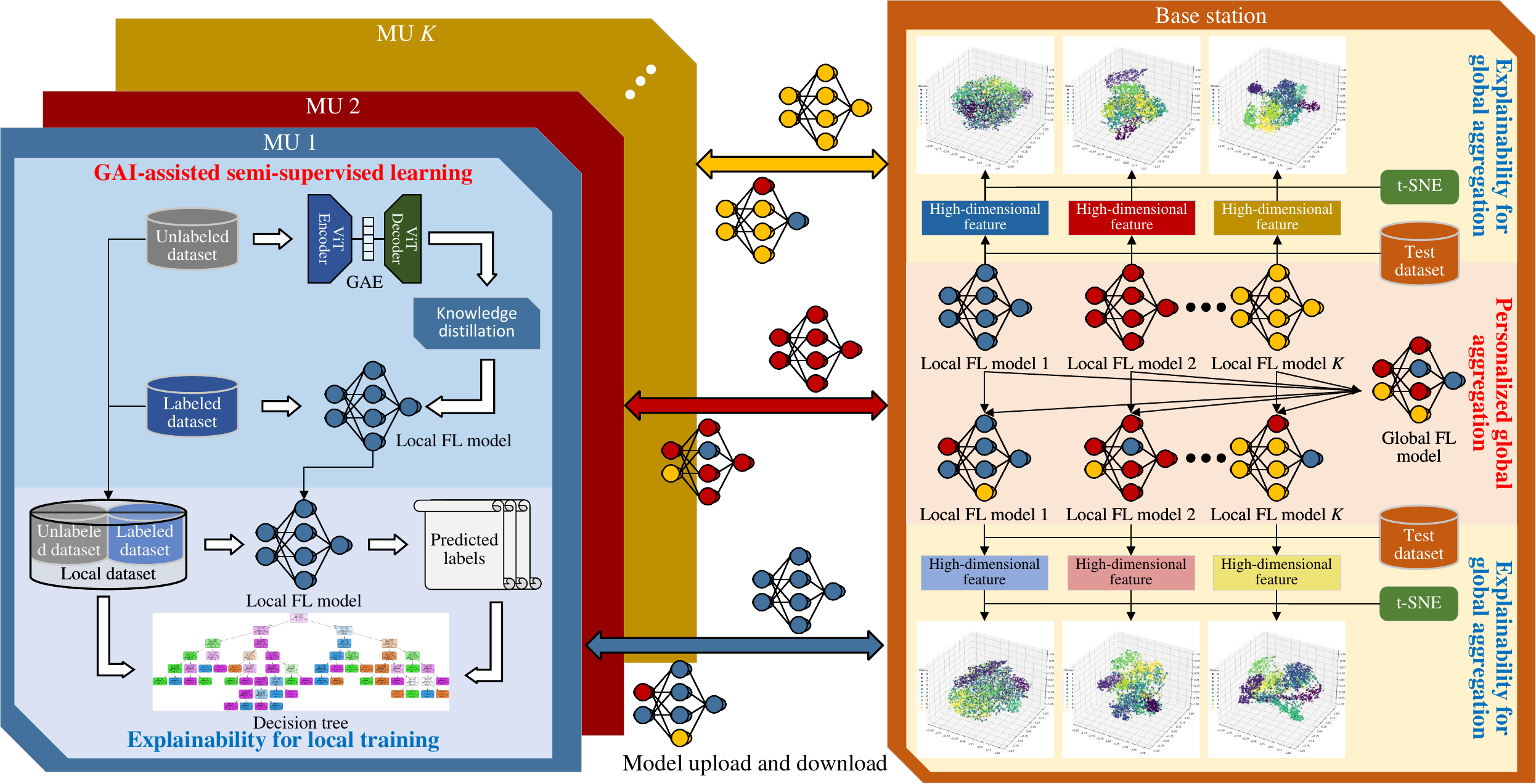}
	\caption{The illustration of the proposed XPFL framework.}
	\label{fig:XPFL}
\end{figure*}

\subsection{Overview of the XPFL Framework}
This subsection presents the implementation overview of the XPFL framework. Assuming that XPFL begins at a round $t$, as depicted in \textbf{Fig. \ref{fig:XPFL}}, the following steps summarize the process of the XPFL framework: 

\subsubsection{GAI-assisted semi-supervised learning} 
To take full advantage of local data set $\mathcal{D}_k$ on the MU, including the labeled data set $\mathcal{D}_k^\mathrm{L}$ and the unlabeled data set $\mathcal{D}_k^\mathrm{U}$, 
each MU $k$ adopts the presented GFed algorithm to perform local training. 
Specifically, first, the supervised and unsupervised learning strategies are applied to labeled and unlabeled data, respectively. Then, the knowledge learned from the local data set $\mathcal{D}_k$ is integrated into the local FL model through KD-based semi-supervised learning. More details about GFed are presented in \textbf{Algorithm \ref{alg:GFed1}}.
Thus, the problem of label scarcity could be solved and the learning efficiency of the local FL model could be guaranteed.  

\subsubsection{Explainable local model}
To enhance the explainability of the local FL model, the proposed XFed mechanism employs a DT as the model explainer. The DT is first trained to fit the local FL model such that, for any given input, the outputs of both the DT and the local FL model are identical. Consequently, the DT can serve as an efficient explainer for the predictions of the local FL model due to its white-box nature.  \textbf{Algorithm \ref{alg:XFed1}} describes the implements of the process in detail. Thus, we realize the explainability of the local FL model in local training.

\subsubsection{Personalized global aggregation}
To mitigate the effects of non-IID data across MUs, the GFed algorithm introduces a novel global aggregation process that facilitates parameter sharing among local FL models. Initially, we perform parameter aggregation using the FedAvg algorithm to obtain the global FL model. Each local FL model is then updated by fusing the local and global FL models in a proportion determined by the difference in their weights. 
This process is described in \textbf{Algorithm \ref{alg:GFed2}}.
As a result, each MU receives a unique FL model after the global aggregation, achieving a personalized FL that allows local models to benefit from parameter sharing while retaining their personalized characteristics.

\subsubsection{Visual global aggregation}
To ensure the reliability and effectiveness of global aggregation, explainability is critical, though it is often overlooked. In the XFed mechanism, t-SNE is used to visualize the alignment of the local FL model with the data distribution. By comparing the visualization results before and after the model update, we can directly observe the changes in each local FL model during global aggregation.
This process is described in \textbf{Algorithm \ref{alg:XFed2}}.
This visualization enables human interpreters to assess the effectiveness of the aggregation process.

After global aggregation, the BS transmits each updated local FL model $w_{k,t}$ to the corresponding MU $k$ for preparing the next FL training:
\begin{equation}\label{eq:broadcast}
	w_{k,t+1}=w_{k,t}, k \in \mathcal{K}
\end{equation}
where $w_{k,t+1}$ is the initial local FL model on the MU $k$ in the $t+1$-th communication round.

To aid comprehension, we assume that $T$ denotes the number of communication rounds, $G$ indicates the iterations of local training, and \textbf{Algorithm \ref{alg:XPFL}} outlines the workflow of the XPFL framework.
\begin{algorithm}
	\caption{XPFL framework}
	\label{alg:XPFL}
	\begin{algorithmic}[1]
		\REQUIRE $G$, $T$, $\mathcal{D}_k$.
		\ENSURE $\mathbf{w}_{k,T}$.
		\FOR{$t=1,2,...,T$}
		\STATE{\textbf{Local Training in MU}}
		\FOR{each MU $k$ in $\mathcal{K}$}
		\FOR{$i=1,2,...,G$}
		\STATE{Perform the GFed algorithm to train the local FL model $w_{k,t}$ on $\mathcal{D}_k$ according to  \textbf{Algorithm \ref{alg:GFed1}}}.
		\ENDFOR
		\STATE{Perform the XFed mechanism to explain the local FL model $w_{k,t}$ according to  \textbf{Algorithm \ref{alg:XFed1}}}.
		\ENDFOR
		\STATE{\textbf{Global Aggregation in BS}}
		\STATE{Perform the GFed algorithm to aggregate and update each local FL model $w_{k,t}$ according to \textbf{Algorithm \ref{alg:GFed2}}}.
		\STATE{Perform the XFed mechanism to visualize the local FL model before and after updating according to \textbf{Algorithm \ref{alg:XFed2}}}.
		\STATE{Broadcast each updated local FL model $w_{k,t}$ to the corresponding MU $k$ according to Eq. (\ref{eq:broadcast}).}
		\ENDFOR
	\end{algorithmic}
\end{algorithm}

\subsection{The Proposed GFed Algorithm}
To address the label scarcity and non-IID problems in FL, the GFed algorithm is proposed, which mainly makes improvements in local training and global aggregation, respectively. As shown in \textbf{Fig. \ref{fig:XPFL}}, the description of the GFed algorithm is as follows:

\subsubsection{GAI-assisted semi-supervised learning}
The traditional FL algorithm is based on supervised learning, such as FedAvg, FedSGD \cite{zhang2023federated}, etc., ignoring that most data of the MU is not labeled, which leads to the knowledge hidden in the unlabeled data being lost. 
As a solution, we adopt the GAE to learn the unlabeled data in an unsupervised way while allowing the local FL model still learn the labeled data in a supervised way. Thereafter, we present a KD-based semi-supervised learning to enable the local FL model to learn the knowledge from the labeled data and the GAE simultaneously.

\begin{figure}[htbp]
	\centering
		\includegraphics[width=8.5cm]{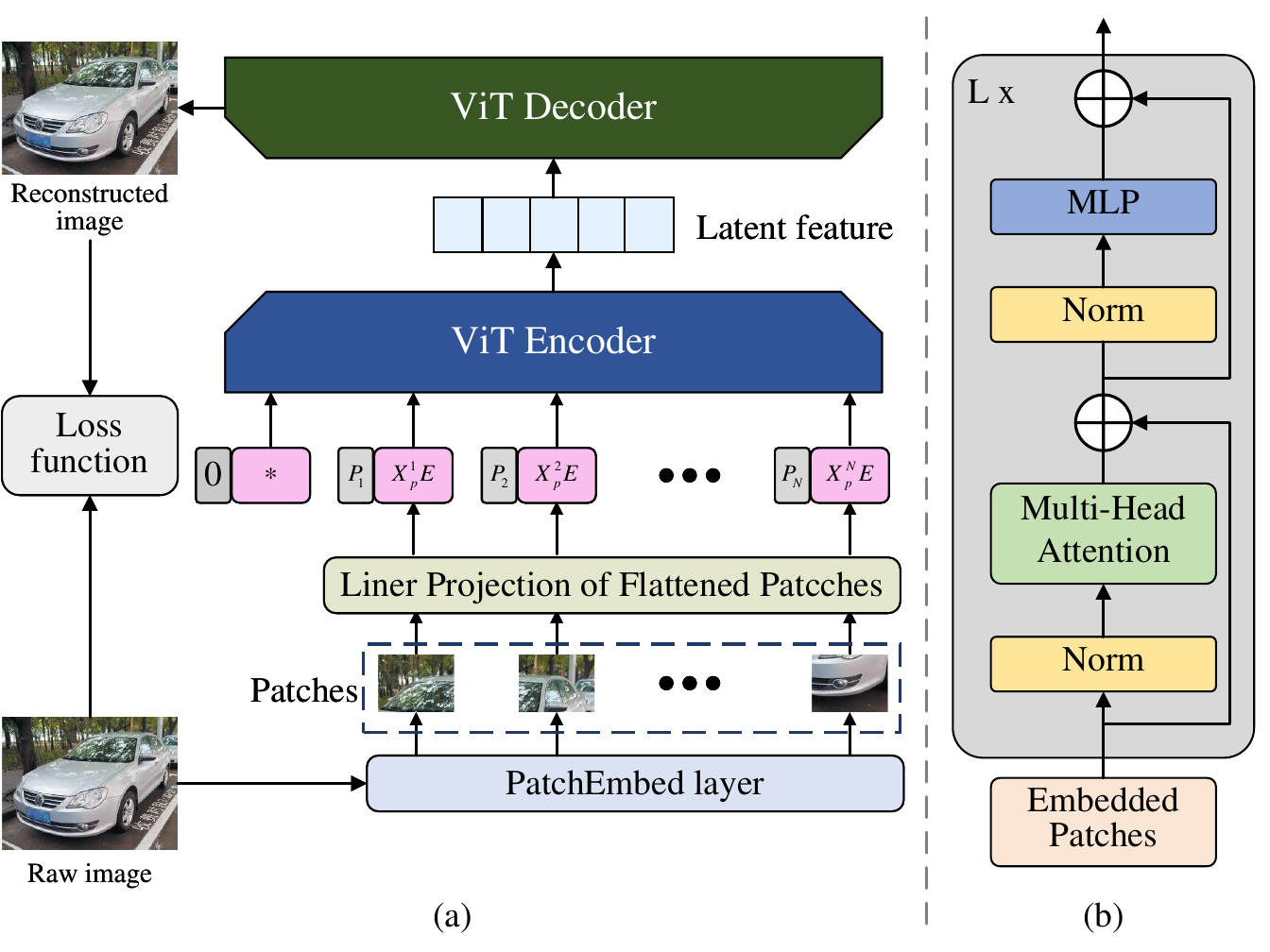}
		%\caption{fig1}

	\caption{The illustration of GAE. (a) The unsupervised learning is based on GAE. (b) The core architecture of ViT.}
	\label{fig:GAE}
\end{figure}

\textbf{Unlabeled knowledge extraction via GAE}: Since we consider disaster monitor scenarios, the main data format is the image, and thus we construct the GAE model based on the vision transformer (ViT).
Compared to traditional convolutional neural networks (CNNs), ViT has displayed superior feature analysis capabilities in various visual tasks, such as image classification, object detection, and feature extraction \cite{jiang2024large3d}. Therefore, we employ ViT as the encoder and decoder of the GAE, integrating the feature extraction capabilities of ViT with the unsupervised feature reconstruction capabilities of the autoencoder.
As shown in Fig. \ref{fig:GAE}(a), the process of knowledge acquisition from unlabeled data based on GAE is as follows:

First, we denote \(H\), \(W\), and \(C\) as the height, width, and channels of the image sample $x_{k,i}^\mathrm{U}$, respectively. We input $x_{k,i}^\mathrm{U}$ into a PatchEmbed layer, and $x_{k,i}^\mathrm{U}$ is converted into \(N\) patches of size \((P^2 \times C)\), where \(P^2\) represents the number of segments into which the image is divided. Hence, \(N = \frac{W \times H}{P^2}\). 

Then, the sequence \(X_p\) composed of these \(N\) patches undergoes the Patch Embedding operation \cite{dosovitskiy2020image}. Specifically, each patch in \(X_p\) bypasses a linear transformation, reducing the dimensionality of the sequence to \(D\) and resulting in a linear embedding sequence:
\begin{equation}\label{eq:GFed1}
	\mathbf{Z}_0 = \left[X_p^1E; X_p^2E; \ldots; X_p^N E\right],
\end{equation}
where \(E\) is the linear transformation (i.e., a fully connected layer) with input dimensions \((P^2 \times C)\) and output dimensions \(N\). \(X_p^i\) represents the \(i\)-th patch in \(X_p\).

Next, the position vectors from the positional encoding, which contain positional information, are linearly combined with $\mathbf{Z}_0$ to obtain the input sequence of ViT as follows:
\begin{equation}\label{eq:GFed2}
	\mathbf{Z} = \left[X_p^1E + P_1; X_p^2E + P_2; \ldots; X_p^N E + P_N\right],
\end{equation}
where \(P_i\) denotes the \(i\)-th position vector.

Subsequently, the encoder of the GAE extracts features from \(\mathbf{Z}\):
\begin{equation}\label{eq:GFed3}
	F_i = \operatorname{ViTEncoder}(\mathbf{Z}),
\end{equation}
where $F_i$ is the latent feature of the raw data $x_{k,i}^\mathrm{U}$ and $\operatorname{ViTEncoder}(\cdot)$ represents the encoder based on ViT. 
Fig. \ref{fig:GAE} displays the core architecture of ViT, with its core being the multi-head attention layer. ViT can learn the relationships between pixels through the multi-head attention mechanism, enabling more accurate feature extraction and realistic image reconstruction. 
The multi-head attention layer is essentially composed of multiple self-attention heads concatenated together. A self-attention head is derived from a single self-attention layer, which can be calculated by:
\begin{equation}\label{eq:GFed4}
	head_i = \operatorname{Attention}(\mathbf{Q}, \mathbf{K}, \mathbf{V}) = \operatorname{softmax} \left( \frac{\mathbf{QK}^T}{\sqrt{d_k}} \right) \mathbf{V}, 
\end{equation}
where \(\mathbf{Q}\) is the query vector, \(\mathbf{K}\) is the matching vector corresponding to \(\mathbf{Q}\), and \(\mathbf{V}\) is the information vector. \(\mathbf{Q}\), \(\mathbf{K}\), and \(\mathbf{V}\) are all obtained through linear transformations of the input \(\mathbf{Z}\), i.e., \(\mathbf{Q} = \mathbf{Z}\mathbf{W}_\mathbf{Q}\), \(\mathbf{K} = \mathbf{Z}\mathbf{W}_\mathbf{K}\), and \(\mathbf{V} = \mathbf{Z}\mathbf{W}_\mathbf{V}\), where \(\mathbf{W}_\mathbf{Q}\), \(\mathbf{W}_\mathbf{K}\), and \(\mathbf{W}_\mathbf{V}\) are the respective weight matrices. \(d_k\) is the scaling factor.
Assuming the number of heads is \( h \), the multi-head self-attention can be calculated by:
\begin{equation}\label{eq:GFed5}
	\begin{split}
		&\operatorname{MultiheadAttention}(\mathbf{Q}, \mathbf{K}, \mathbf{V}) =\\
		&\text{concat}(head_1, \cdots, head_h)\mathbf{W}_\text{mha},
	\end{split}
\end{equation}
where \(\mathbf{W}_\text{mha}\) represents the weights of the multi-head attention layer.

Finally, the decoder of the GAE performs data reconstruction based on the latent feature $F_i$:
\begin{equation}\label{eq:GFed6}
\hat{x}_{k,i}^\mathrm{U} = \operatorname{ViTDecoder}(F_i),
\end{equation}
where $\hat{x}_{k,i}^\mathrm{U}$ represents the reconstructed data and $\operatorname{ViTDecoder}(\cdot)$ is the decoder based on the ViT. 
The mean-square error is used as the loss function of the GAE and thus captures the reconstruction error, which can be given by:
\begin{equation}\label{eq:GFed7}
	\mathcal{L}^\mathrm{U}=\frac{1}{D_k^\mathrm{U}}\sum\limits_{i=1}^{D_k^\mathrm{U}}(x_{k,i}^\mathrm{U}-\hat{x}_{k,i}^\mathrm{U})^2,
\end{equation}
where $D_k^\mathrm{U}$ is the size of the unlabeled data set $\mathcal{D}_k^\mathrm{U}$. The GAE aims to minimize $\mathcal{L}^\mathrm{U}$ to obtain the better feature representation of the unlabeled data, namely the $F_i$. 

\textbf{Semi-supervised learning based on KD}: After obtaining the latent knowledge of the unlabeled data set $\mathcal{D}_k^\mathrm{U}$ through GAE, the local FL model performs KD-based semi-supervised learning on the labeled data set $\mathcal{D}_k^\mathrm{L}$. Specifically, we assume the local FL model is constructed based on the CNNs, and it learns the knowledge of $\mathcal{D}_k^\mathrm{L}$ via supervised learning. Simultaneously, we employ KD to transfer the knowledge learned by the GAE to the local FL model. 
KD is a popular transfer learning technique that involves a pre-trained teacher model and a non-pretrained student model. During the local training, the goal of KD is to transfer the knowledge from the teacher model to the student model \cite{shen2020federated}. Here, we treat the local FL model as the student model and the pre-trained GAE as the teacher model.

We consider the loss of supervised learning when the local FL model training on $\mathcal{D}_k^\mathrm{L}$, namely compute the error between the output of models and labels. Assume we consider a classification task and use the cross-entropy as the supervised learning loss function:
\begin{equation}\label{eq:GFed8}
	\mathcal{L}_\mathrm{CE}(\mathbf{y}, \mathbf{\hat{y}})=-\sum_{i=1}^{M} y_{i} \log \left(\hat{y}_{i}\right),
\end{equation}
where $\mathbf{y}=[y_1,y_2,..., y_M]$ represents the labels, $y_i=1$ means the input data $x_{k,i}^\mathrm{L}$ belongs to the $i$-th class, otherwise 0; $\mathbf{\hat{y}}=[\hat{y}_1,\hat{y}_2,..., \hat{y}_M]$ represents the predicted probabilities, $\hat{y}_i$ is the probability predicted as the $i$-th class; $M$ is the total number of categories.

During the training of the local FL model in a supervised way, the GAE as the teacher model regards its data feature $F_i$ extracted from the input data $x_{k,i}^\mathrm{L}$ as the soft labels, directing the local FL model to learn the ability of feature representation  \cite{aguilar2020knowledge}. We denote the data features extracted from $x_{k,i}^\mathrm{L}$ of the local FL model and GAE are denoted as $F_i^\mathrm{t}$ and $F_i^\mathrm{s}$, respectively. Then, the KD loss of the local FL models can be expressed as:
\begin{equation}\label{eq:GFed11}
	\mathcal{L}_\mathrm{KD}=\operatorname{KL}(\operatorname{softmax}(F_i^\mathrm{t})||\operatorname{softmax}(F_i^\mathrm{s}))/\mathcal{L}^\mathrm{U},
\end{equation}
where ${\operatorname{KL}}(\cdot)$ means the Kullback–Leibler (KL) divergence; ${\operatorname{softmax}}(\cdot)$ function could be expressed as: $\operatorname{softmax}\left(z,\tau \right)=\frac{\exp \left(z / \tau\right)}{\sum_j \exp \left(z_j / \tau\right)}$, where $z$ is the input and $\tau$ is a temperature factor to control the importance of each soft label \cite{cho2019efficacy}. 
Considering incorrect predictions from the teacher model may mislead the student model in the training, we introduce the training quality of GAE, namely $\mathcal{L}^\mathrm{U}$, to adjust the distillation loss, adaptively. 

The semi-supervised learning loss of the local FL model can be expressed as:
\begin{equation}\label{eq:GFed12}
	\mathcal{L}^\mathrm{S} = \mathcal{L}_\mathrm{CE} + \lambda\mathcal{L}_\mathrm{KD},
\end{equation}
where $\lambda$ is an adjust factor. 

By the GAI-assisted semi-supervised learning, we can fully exploit the data set $\mathcal{D}_k^\mathrm{L}$ and $\mathcal{D}_k^\mathrm{U}$ to train the local FL model, and thus address the label scarcity.

\subsubsection{Personalized global aggregation}
The typical FL objective is to minimize the aggregation of local FL losses, resulting in a common output for all MUs using a global FL model without any personalization. However, the performance of the global FL model may decline when faced with heterogeneous local data distributions across individual MUs \cite{tan2022towards}. As a remedy, personalized FL is adopted in this paper.

Since the semi-supervised local training, the unlabeled data set is also fully utilized, hence the aggregation formulation Eq. (\ref{eq:aggregation}) could be rewritten as:
\begin{equation}\label{eq:aggregation3}
	w_{g,t}=\frac{\sum\limits_{k=1}^{K} \rho_k w_{k,t} D_k C(w_{k,t})}{\sum\limits_{k=1}^{K} \rho_k D_k C(w_{k,t})},
\end{equation}
where $\rho_k=D_k^\text{L}/D_k^\text{U}$ represents the labeled and unlabeled data volume ratio. The MU $k$ owns more labeled data means the corresponding local FL model $w_{k,t}$ is more reliable, hence a larger weight $\rho_k$ is given when aggregation. Thus, we ensure the aggregation is reliable under the label scarcity condition.

Our goals are to construct personalized models by modifying the FL model aggregation process. 
In particular, we use the cosine distance to measure the similarity between the local and global FL models, which could be given as:
\begin{equation}\label{eq:GFed13}
	\psi_{k,t}=(1+\frac{\mathbf{w}_{k,t}\cdot\mathbf{w}_{g,t}}{\rVert\mathbf{w}_{k,t}\rVert\cdot\rVert\mathbf{w}_{g,t}\rVert})/2,
\end{equation}
where $\mathbf{w}_{k,t}$ and $\mathbf{w}_{g,t}$ represent the parameters vector of the local and global FL models, respectively. Note $\psi_{k,t}\in[0,1]$, and the larger $\psi_{k,t}$ means the more similar $w_{k,t}$ and $w_{g,t}$ are.
After global aggregation, each local FL model $w_{k,t}$ fuses with the global FL model $w_{g,t}$ according to $\psi_{k,t}$, which could be expressed as:
\begin{equation}\label{eq:GFed14}
	w_{k,t} = \psi_{k,t}\cdot w_{k,t} + (1-\psi_{k,t}) w_{g,t}.
\end{equation}

Eq. (\ref{eq:GFed14}) illustrates our usage of model-based FL approaches, aimed at achieving global model personalization. This approach enhances the adaptation performance of local FL models. 
We have summarized the proposed GFed algorithm in \textbf{Algorithm \ref{alg:GFed1}} and \textbf{Algorithm \ref{alg:GFed2}}. 
\begin{algorithm}
	\caption{GFed in local training}
	\label{alg:GFed1}
	\begin{algorithmic}[1]
		\REQUIRE $\mathcal{D}_k$, $w_{k,t}$.
		\ENSURE $w_{k,t}$.
		\FOR{$x_{k,i}^\mathrm{U} \in \mathcal{D}_k^\text{U}$}
		\STATE{Obtain the reconstructed data $\hat{x}_{k,i}^\mathrm{U}$ according to Eqs. (\ref{eq:GFed1}) - (\ref{eq:GFed6}).}
		\STATE{Update GAE by minimizing Eq. (\ref{eq:GFed7}).}
		\ENDFOR
		\FOR{$x_{k,i}^\mathrm{L} \in \mathcal{D}_k^\text{L}$}
		\STATE{Calculate the supervised learning loss according to Eq. (\ref{eq:GFed8}).}
		\STATE{Calculate the KD loss according to Eq. (\ref{eq:GFed11}).}
		\STATE{Update the local FL model $w_{k,t}$ by minimizing semi-supervised learning loss in Eq. (\ref{eq:GFed12}).}
		\ENDFOR
	\end{algorithmic}
\end{algorithm}
\begin{algorithm}
	\caption{GFed in global aggregation}
	\label{alg:GFed2}
	\begin{algorithmic}[1]
		\REQUIRE $w_{k,t}$.
		\ENSURE $w_{k,t}$.
		\STATE{Obtain the global FL model $w_{g,t}$ according to Eq. (\ref{eq:aggregation3}).}
		\FOR{each local FL model $w_{k,t}$}
		\STATE{Calculate the cosine distance between the local and global FL models according to Eq. (\ref{eq:GFed13}).}
		\STATE{Update local FL model $w_{k,t}$ according to Eq. (\ref{eq:GFed14}).}
		\ENDFOR
	\end{algorithmic}
\end{algorithm}

With the implementation of the GFed algorithm, on the one hand, by combining the advantages of the GAI and KD, we utilize semi-supervised learning to make the unlabeled data fully utilized and solve the label scarcity. On the other hand, we develop a cosine distance-based personalized global aggregation to allow local FL models to benefit from the global aggregation and adapt to the respective local data set, thus overcoming the effect of non-IID data in FL.

\subsection{The Proposed XFed Mechanism}
To achieve the explainability of FL in local training and global aggregation phases, we present the XFed mechanism. As illustrated in \textbf{Fig. \ref{fig:XPFL}}, the description of the XFed mechanism is described as follows:

\subsubsection{Explainable local model}
Following local training, the local FL model is deployed to perform real-world tasks and make decisions, where the explainability of its predictions is crucial, especially in fields related to life, health, and safety. Therefore, it is essential to incorporate a white-box model as an explainer for the local FL model.
In this paper, we focus on image data as the primary format for training on each MU, which often involves non-linear features. Additionally, each MU operates with limited computational resources, making prolonged training infeasible. DT, a typical white-box model, is capable of handling non-linear features efficiently while requiring minimal computational resources  \cite{dabiri2022applications}. This makes it suitable for resource-limited MUs. Therefore, we select DT as the explainer to interpret the relationship between the input and the predictions of the local FL model.

The DT adopts a tree structure for sample classification, where the leaves indicate the class labels and the branches are the attributes (or conditions) \cite{dabiri2022applications}.
Note that the attributes are not selected randomly but through a factor named information gain.
We expect that DT could as an explainer of the local FL model, and the goal of DT is to model the data set that consists of the input and output of the local FL model.
Hence, we define the training data set of DT as $\hat{\mathcal{D}}_k=\{(x_{k,i},\hat{y}_{k,i})\}_{i=1}^{D_k}$, where $x_{k,i}$ represents the $i$-th data sample in $\mathcal{D}_k$ and $\hat{y}_{k,i}$ denotes the predicted label of the local FL model according to $x_{k,i}$. The entropy of $\hat{\mathcal{D}}_k$ in DT could be expressed as:
\begin{equation}\label{eq:XFed1}
	\operatorname{e}(\hat{\mathcal{D}}_k)= -\sum_{i=1}^{M} p_i \log \left(p_i\right),
\end{equation}
where $p_i$ represents the probability that the $i$-th category appears in $\hat{\mathcal{D}}_k$. 
Suppose $\{\hat{\mathcal{D}}_{k,1}, \hat{\mathcal{D}}_{k,2},\ldots, \hat{\mathcal{D}}_{k,N_A}\}$ are subsets of $\hat{\mathcal{D}}_k$ divided through all possible values of a certain attribute $A$, where $N_A$ is the number of subsets. The expected information can be calculated as:
\begin{equation}\label{eq:XFed2}
	\operatorname{e}_A(\hat{\mathcal{D}}_k)=\sum_{j=1}^{N} \frac{\left|\hat{\mathcal{D}}_{k,j}\right|}{D_k} \operatorname{e}\left(\hat{\mathcal{D}}_{k,j}\right),
\end{equation}
where $|\cdot|$ represents the size of a data set. The information gain is the difference between Eq. (\ref{eq:XFed1}) and Eq. (\ref{eq:XFed2}), which could be given by:
\begin{equation}\label{eq:XFed3}
	\operatorname{I}_A(\hat{\mathcal{D}}_k)= \operatorname{e}(\hat{\mathcal{D}}_k)-\operatorname{e}_A(\hat{\mathcal{D}}_k).
\end{equation}

Then, the information gain ratio can be calculated by \cite{ruggieri2002efficient}:
\begin{equation}\label{eq:XFed3-1}
	\operatorname{G}_A(\hat{\mathcal{D}}_k)=\frac{\operatorname{I}_A(\hat{\mathcal{D}}_k)}{\operatorname{S}_A(\hat{\mathcal{D}}_k)},
\end{equation}
where $\operatorname{S}_A(\hat{\mathcal{D}}_k)=-\sum_{j=1}^{N}\frac{\left|\hat{\mathcal{D}}_{k,j}\right|}{D_k}\log_{2}\frac{\left|\hat{\mathcal{D}}_{k,j}\right|}{D_k}$ represents the split entropy of the attribute $A$.
Thus, DT selects the attribute with the most significant information gain ratio as the split attribute. Each split will cause the tree to grow one level taller and a DT is constructed eventually, which means the DT finishes the fit to the training data set  $\hat{\mathcal{D}}_k$. The constructed DT is used as the explainer of the local FL model $w_{k,t}$ and denotes as $DT_{k,t}$.
\subsubsection{Visual global aggregation}
In global aggregation, each local FL model shares the parameter information to obtain the global FL model and then performs model fusion according to Eq. (\ref{eq:aggregation3}). 
To ensure this aggregation is effective, we need to explain whether the feature extraction ability of the FL model improves before and after the update. 
As a solution, we employ t-SNE, a technique that visualizes high-dimensional data by assigning each data point a location on a two or three-dimensional map. This enables visualization of how each local FL model fits the distribution of data. More importantly, in the process of visualization, t-SNE can preserve the characteristics of these high-dimensional data \cite{liu2019sequence}, ensuring the reliability of explainability.

The t-SNE algorithm seeks to map both high-dimensional data points and their corresponding low-dimensional analogs in a way that preserves the similarity of the data points \cite{liu2019sequence}. The optimal low-dimensional analogs are obtained by minimizing KL divergence.
To enable the explainability of global aggregation through t-SNE, the following steps are taken:

First, we suppose there is a test data set $\mathcal{D}_\text{test}=\{(x_i^\text{t},y_i^\text{t})\}_{i=1}^{D_\text{test}}$ on the BS, where $x_i^\text{t}$ is the $i$-th sample, $y_i^\text{t}$ is the corresponding label, and $D_\text{test}$ represents the size of the test data set. % Note that $D_\text{test}\ll D_k, \forall k\in\mathcal{K}$. 
Each sample $x_i^\text{t}$ is fed into the local FL model $w_{k,t}$, and the corresponding feature representation $F_i^k$ is obtained according to Eq. (\ref{eq:GFed6}). We assume the local FL model $w_{k,t}$ is not yet updated by the personalized global aggregation.
All the feature reputation compose a high-dimensional data set $\mathcal{H}_{k}=\{F_i^k\}_{i=1}^{D_\text{test}}$, which is used as the input of t-SNE.

Then, the probability of similarity between two data points, $F_i^k$ and $F_j^k$, can be given by \cite{van2008visualizing}:
\begin{equation}\label{eq:XFed4}
	p_{i \mid j}=\frac{\exp \left(-\left\|F_j^k-F_i^k\right\|^{2} / 2 \sigma_{j}^{2}\right)}{\sum_{n \neq j}^{D_\text{test}} \exp \left(-\left\|F_j^k-F_n^k\right\|^{2} / 2 \sigma_{j}^{2}\right)},
\end{equation}
\begin{equation}\label{eq:XFed5}
	p_{j \mid i}=\frac{\exp \left(-\left\|F_i^k-F_j^k\right\|^{2} / 2 \sigma_{i}^{2}\right)}{\sum_{n \neq i}^{D_\text{test}} \exp \left(-\left\|F_i^k-F_n^k\right\|^{2} / 2 \sigma_{i}^{2}\right)},
\end{equation}
where $\sigma_{i}$ and $\sigma_{j}$ are the vector variance of the Gaussian function centered on the data point $F_i^k$ and $F_j^k$, respectively. 
Thus, in the Gaussian space, the joint distribution of any two data points can be expressed as:
\begin{equation}\label{eq:XFed6}
	p_{ij}=\frac{p_{i\mid j}+p_{j\mid i}}{2D_\text{test}}.
\end{equation}

The $t$ distribution is used to sample in low-dimensional space to obtain the low-dimensional analog data set $\mathcal{A}_k=\{\hat{F}_i^k\}_{i=1}^{D_\text{test}}$. 
Similarly, the joint distribution between the analog data points can be given by \cite{van2008visualizing}:
\begin{equation}\label{eq:XFed7}
	q_{ij}=\frac{ \left(1+\left\|\hat{F}_i^k-\hat{F}_j^k\right\|^{2}\right)^{-1}}{\sum_{n \neq l}  \left(1+\left\|\hat{F}_n^k-\hat{F}_l^k\right\|^{2}\right)^{-1}}.
\end{equation}

After obtaining the low-dimensional analog data points, we use KL divergence to measure the accuracy of the simulation between these analogs and the high-dimensional data points. The expression for KL divergence is as follows:
\begin{equation}\label{eq:XFed8}
	C=\operatorname{KL}(P||Q)=\sum_{i=1}^{D_\text{test}}\sum_{j=1}^{D_\text{test}}p_{ij}\log_{2}\frac{p_{ij}}{q_{ij}}.
\end{equation}

To maximize the accuracy of simulation, the gradient descent method is used to optimize the result of KL divergence in Eq. (\ref{eq:XFed8}), which could be expressed as:
\begin{equation}\label{eq:XFed9}
	\frac{\partial C}{\partial \hat{F}_i^k}=4\sum_j^{D_\text{test}}(p_{ij}-q_{ij})(\hat{F}_i^k-\hat{F}_j^k)(1+\rVert\hat{F}_i^k-\hat{F}_j^k\rVert^2)^{-1}.
\end{equation}

Finally, based on the above calculations and iterating multiple times $\varGamma$, we obtain the optimal low-dimensional analog data points $\mathcal{F}_k^\mathrm{l}$. Thus, we can visualize the feature representation extracted by the local FL $w_{k,t}$ from the test data set $\mathcal{D}_\text{test}$, which could reflect the quality of the model indirectly. 
Similarly, for the updated local FL model, we can get the corresponding low-dimensional analog data points $\widetilde{\mathcal{F}}_k^\mathrm{l}$ and visualization results.
By observing the difference between the visualization results before and after the model update, we can easily explain whether the model performance is better and identify whether this global aggregation is effective.

\begin{algorithm}
	\caption{XFed in local training}
	\label{alg:XFed1}
	\begin{algorithmic}[1]
		\REQUIRE $\mathcal{D}_k$, $w_{k,t}$.
		\ENSURE $DT_{k,t}$.
		\STATE{Obtain the DT training data set $\hat{\mathcal{D}}_k$ by feeding $\mathcal{D}_\text{k}$ into $w_{k,t}$.}
		\STATE{Calculate the entropy of $\hat{\mathcal{D}}_k$ according to Eq. (\ref{eq:XFed1}).}
		\STATE{Initial the DT $DT_{k,t}$.}
		\WHILE{$DT_{k,t}$ can be spilted}
		\FOR{each attribute $A$}
		\STATE{Divide $\hat{\mathcal{D}}_k$ according to the attribute $A$ and obtain the subsets $\{\hat{\mathcal{D}}_{k,1}, \hat{\mathcal{D}}_{k,2},\ldots, \hat{\mathcal{D}}_{k,N_A}\}$.}
		\STATE{Calculate the expected information $\operatorname{e}_A(\hat{\mathcal{D}}_k)$ according to Eq. (\ref{eq:XFed2}).}
		\STATE{Calculate the information gain $\operatorname{I}_A(\hat{\mathcal{D}}_k)$ according to Eq. (\ref{eq:XFed3}).}
		\STATE{Calculate the information gain ratio $\operatorname{G}_A(\hat{\mathcal{D}}_k)$ according to Eq. (\ref{eq:XFed3-1})}
		\ENDFOR
		\STATE{$DT_{k,t}$ splits according to the split attribute that has the largest information gain ratio.}
		\ENDWHILE
	\end{algorithmic}
\end{algorithm}
\begin{algorithm}
	\caption{XFed in global aggregation}
	\label{alg:XFed2}
	\begin{algorithmic}[1]
		\REQUIRE $\mathcal{D}_\text{test}$, $w_{k,t}$, $\varGamma$.
		\ENSURE $\mathcal{F}_k^\mathrm{l}$.
		\STATE{Obtain the high-dimensional data set $\mathcal{F}_k^\mathrm{h}$ by feeding $\mathcal{D}_\text{test}$ into $w_{k,t}$.}
		\STATE{Calculate pairwise affinities $p_{ij}$ in high-dimensional space according to Eqs. (\ref{eq:XFed4}) - (\ref{eq:XFed6}).}
		\STATE{Initial analog data set $\mathcal{F}_k^\mathrm{l}$.}
		\FOR{each iteration in $\varGamma$}
		\STATE{Calculate pairwise affinities $q_{ij}$ in low-dimensional space according to Eq. (\ref{eq:XFed7}).}
		\STATE{Calculate the KL divergence between $p_{ij}$ and $q_{ij}$ according to Eq. (\ref{eq:XFed8}).}
		\STATE{Calculate the gradient according to Eq. (\ref{eq:XFed9}) and optimize with the gradient descent method.}
		\ENDFOR
	\end{algorithmic}	
\end{algorithm}

For better understanding, we have summarized the XFed mechanism in \textbf{Algorithm \ref{alg:XFed1}} and \textbf{Algorithm \ref{alg:XFed2}}. 
With the implementation of the XFed mechanism, we first use the white-box model DT as the specific explainer to the local FL model, ensuring the predictions of the local FL model can be explained.
Then, the t-SNE-based explainability method makes the changes of the local FL models during the global aggregation can be captured in a visualized way, thus improving the transparency and trustworthiness of the global aggregation.

\subsubsection{Quantify metric for explainability}
The FL explainability is a highly abstract concept representing the model's transparency and users' trust. Hence, to quantify the explainability of the local FL model of the MU $k$, we explore a new metric, Quality of eXplainability (QoX), which can be expressed as follows:
\begin{equation}\label{eq:QoX}
	QoX_{k}=\frac{Acc(DT_{k,t})D_k}{|w_{k,t}|}+\chi\frac{(C-C^\prime)D_\text{test}}{|w_{k,t}|}, 
\end{equation}
where $Acc(DT_{k,t})$ represents the accuracy of the DT $DT_{k,t}$; $|w_{k,t}|$ denotes the size of the local FL model $w_{k,t}$; $C$ and $C^\prime$ are the KL divergence results of the t-SNE computed on the local FL model before and after the update; $\chi$ is a coefficient factor. 

In Eq. (\ref{eq:QoX}), $Acc(DT_{k,t})D_k$ reflects the decision explainability of the local FL model in the MU and it is larger means the DT as the explainer of the local FL model is more reliable, $(C-C^\prime)D_\text{test}$ represents the aggregation explainability, and it is larger means the improvement of the local FL model is higher before and after aggregation. 
With the improvements of the above two, the FL explainability is increased.
While $|w_{k,t}|$ is larger means the local FL model is more complex and the FL explainability is decreased.

\section{Simulation and Discussion}

\subsection{Simulation Settings}
Our simulations use MNIST \cite{lecun1989handwritten}, Fashion-MNIST \cite{xiao2017fashion}, and CIFAR-10 \cite{krizhevsky2009learning} as evaluation datasets, and they are split in a non-IID way. Specifically, we apply the Dirichlet distribution to generate the non-IID data partition among MUs \cite{li2021model}, where the concentration parameter $\eta$ of the Dirichlet distribution is set to 0.5 by default. As a result, each MU gets relatively few (even no) data samples of some classes. 
We denote $\gamma$ as the proportion of labeled data on each MU and the default is 0.1.
%All experiments are repeated 10 times to avoid randomness, and multiple experimental results are obtained. We average all experimental results as the final results.

\iffalse
The descriptions of the three datasets are as follows:
\begin{itemize}
	\item MNIST \cite{lecun1989handwritten}: It has a training set containing 60,000 samples and a testing set containing 10,000 samples. Each sample is a 28 $\times$ 28 image of handwritten digits associated with a label from 10 classes.
	\item Fashion-MNIST \cite{xiao2017fashion}: It consists of a training set of 60,000 samples and a testing set of 10,000 samples. Each sample is a 28 $\times$ 28 grayscale image associated with a label from 10 classes.
	\item CIFAR-10 \cite{krizhevsky2009learning}: It consists of a training set of 50,000 samples and a testing set of 10,000 samples. Each sample is a 32 $\times$ 32 RGB colour image associated with a label from 10 classes.
\end{itemize}
\fi

The network structures of the FL model and the GAE model are designed as follows:
\begin{itemize}
	\item FL model:
	The architecture of this model is composed of two blocks, each consisting of a convolutional layer and a pooling layer. Additionally, there is a fully connected layer in the model.
	To introduce non-linearity to the model, all convolutional layers are followed by a Rectified Linear Unit (ReLU) activation function.
	\item GAE model:
	Masked Autoencoders (MAE) \cite{he2022masked} is a well-known generative model, which utilizes ViT as the encoder-decoder to learn accurate feature representation of images and perform high-quality construction. Therefore, we adopt the MAE as the GAE model.
\end{itemize}

We consider that there are 10 MUs for FL training.
The number of communication rounds and iterations of the local training are set as $T=40$ and $G=1$, respectively. Our simulations are performed with the PyTorch framework on a server, which has an Intel Xeon CPU (2.4 GHz, 128 GB RAM) and an NVIDIA A800 GPU (80 GB SGRAM). %The code is available at \href{https://github.com/qiyuqianxai/FL-LISC.git}{https://github.com/qiyuqianxai/XPFL.git}. 
\section{Evaluation of the GAE-Based Unsupervised Learning}
To intuitively highlight the advantages of the GAI architecture in learning from unlabeled data, we present a series of restructured images generated by the GAE and the CNN-based autoencoder (CAE). In addition, we apply evaluation metrics such as peak signal-to-noise ratio (PSNR) and Structural similarity index measure (SSIM) to quantify the quality of the reconstructed images. PSNR measures the quality of the reconstructed images, typically expressed in decibels, with higher values indicating better image fidelity. The PSNR is defined as follows:
\begin{equation}
	\mathrm{PSNR}(x,\hat{x}) = 10 \cdot \log_{10} \left( \frac{\mathrm{MAX}_I^2}{\mathrm{MSE}(x,\hat{x})} \right),
\end{equation}
where $x$ and $\hat{x}$ represent the original and reconstructed images, respectively, and $\mathrm{MAX}_I$ refers to the maximum possible pixel value of the image, typically 255 for an 8-bit image.
Similarly, SSIM is a metric that evaluates the perceived similarity between two images, taking into account three components: luminance, contrast, and structure. SSIM is defined as:
\begin{equation}
	\mathrm{SSIM}(x,\hat{x}) = \frac{(2\varphi_{x}\varphi_{\hat{x}} + c_1)(2\phi_{x\hat{x}} + c_2)}{(\varphi_{x}^2 + \varphi_{\hat{x}}^2 + c_1)(\phi_{x}^2 + \phi_{\hat{x}}^2 + c_2)},
\end{equation}
where $\varphi_{x}$ and $\varphi_{\hat{x}}$ denote the means of $x$ and $\hat{x}$; $\phi_{x}^2$ and $\phi_{\hat{x}}^2$ are their variances; $\phi_{x\hat{x}}$ represents their covariance; and $c_1$ and $c_2$ are constants introduced to prevent division by zero.

\begin{figure}[htbp]
	\centering
	\includegraphics[width=8.5cm]{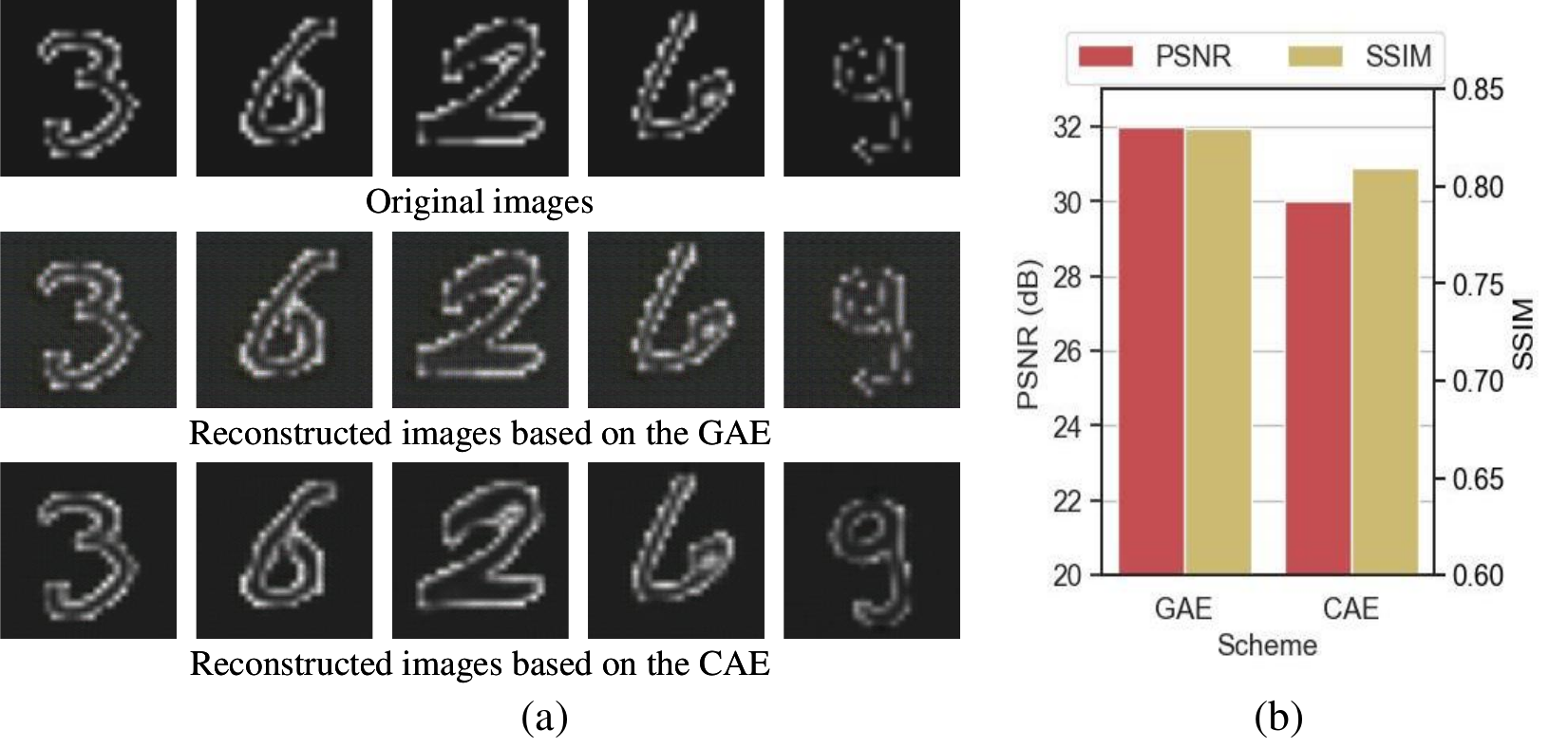}
	\caption{Image transmission results on the MNIST dataset. (a) Visual comparison of image transmission quality. (b) Quantitative comparison of image transmission quality.}
	\label{fig:GSC_exp1}
\end{figure}

\begin{figure}[htbp]
	\centering
	\includegraphics[width=8.5cm]{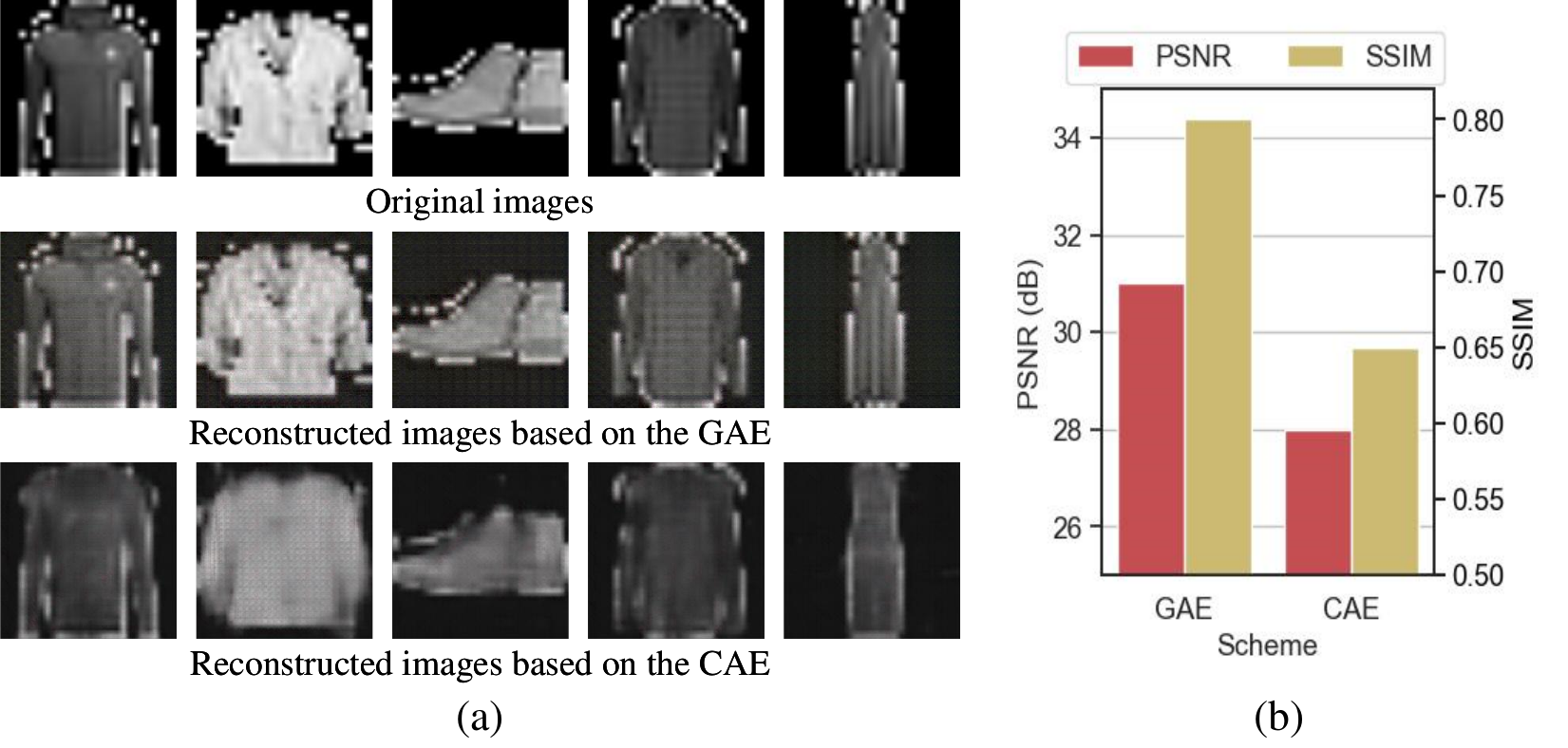}
	\caption{Image transmission results on the Fashion-MNIST dataset. (a) Visual comparison of image transmission quality. (b) Quantitative comparison of image transmission quality.}
	\label{fig:GSC_exp2}
\end{figure}

\begin{figure}[htbp]
	\centering
	\includegraphics[width=8.5cm]{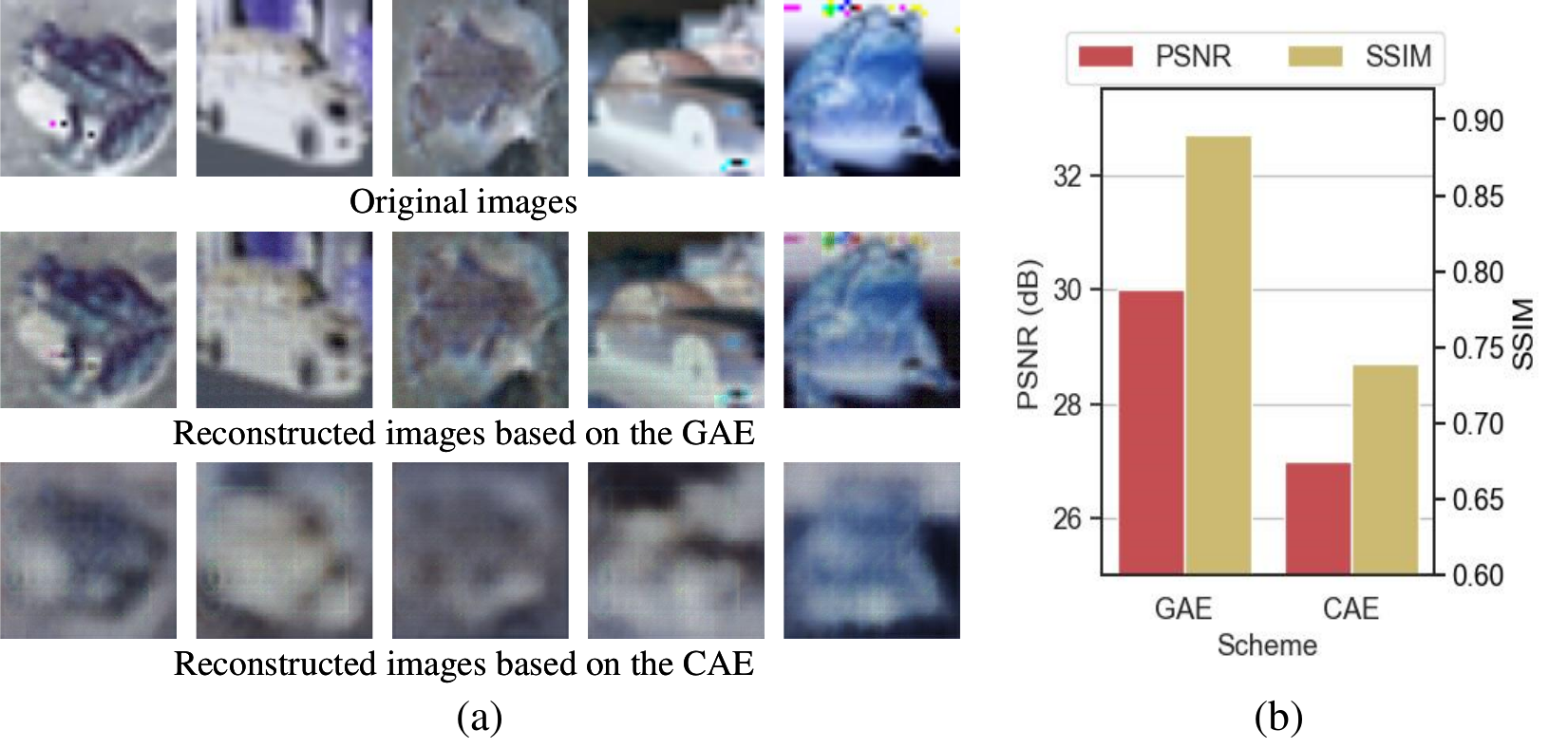}
	\caption{Image transmission results on the CIFAR-10 dataset.  (a) Visual comparison of image transmission quality. (b) Quantitative comparison of image transmission quality.}
	\label{fig:GSC_exp3}
\end{figure}

Fig. \ref{fig:GSC_exp1} presents the evaluation results on the MNIST dataset. As shown in Fig. \ref{fig:GSC_exp1}(a), both the GAE and CAE demonstrate good image transmission quality; however, GAE excels in preserving finer details. The quantitative results in Fig. \ref{fig:GSC_exp1}(b) reveal the differences between the original and reconstructed images in terms of PSNR and SSIM, with GAE achieving higher scores.
Fig. \ref{fig:GSC_exp2} presents the evaluation results on the Fashion-MNIST dataset, where Fig. \ref{fig:GSC_exp2}(a) demonstrates that the images generated by the Generative Autoencoder (GAE) exhibit superior quality. The PSNR and SSIM results in Fig. \ref{fig:GSC_exp2}(b) further confirm the higher quality of the GAE.
Fig. \ref{fig:GSC_exp3} illustrates the evaluation results on the CIFAR-10 dataset. In this case, Fig. \ref{fig:GSC_exp3}(a) indicates that the GAE can accurately reconstruct the original images, whereas the reconstructions produced by the Convolutional Autoencoder (CAE) appear blurry. Additionally, the PSNR and SSIM results in Fig. \ref{fig:GSC_exp3}(b) indicate that the images generated by the GAE are of higher quality than those produced by the CAE.


The superior performance of the GAE can be attributed to the advantages of the GAI architecture (i.e., the MAE), which extracts more precise feature information than CNN architectures. Furthermore, the GAE achieves more accurate image reconstruction in unsupervised learning tasks due to its strong generative capabilities. 

\subsection{Evaluation of XPFL for Label Scarcity}
\begin{figure*}[htbp]
	\centering
	\includegraphics[width=18cm]{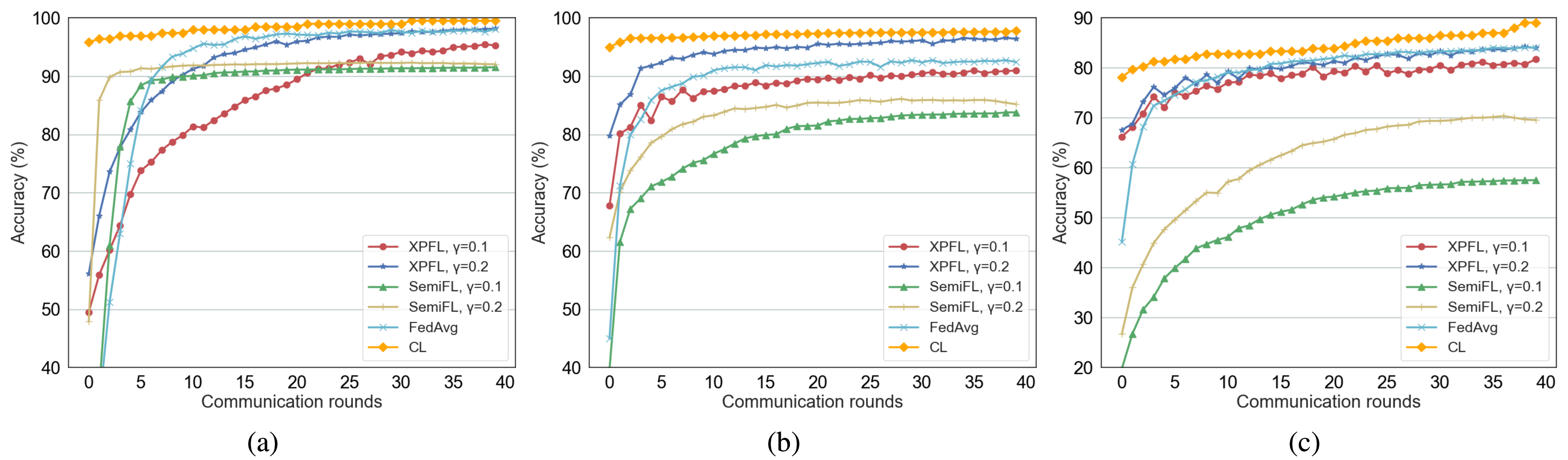}
	\caption{The accuracy of the global FL model under different schemes on datasets (a) MNIST, (b) Fashion-MNIST, and (c) CIFAR-10.}
	\label{fig:exp1}
\end{figure*}

This simulation is presented to evaluate the performance of the XPFL framework under the condition of label scarcity.
In this subsection, the following algorithms are used as the contenders:
\begin{itemize}
	\item SemiFL \cite{diao2022semifl}: This FL approach focuses on solving the challenge of integrating communication-efficient FL with semi-supervised learning.
	\item CL \cite{zhang2018improved}: The model is trained directly using the entire data.
	\item FedAvg \cite{ghimire2022recent}: The common FL approach, which is equivalent to the XPFL without GFed and XFed algorithms.
	\item XPFL: Our proposed novel FL approach, which could solve the problem of label scarcity through semi-supervised local training.
\end{itemize}

Note that all data used by the FedAvg and CL are labeled, while for the data used by the XPFL and SemiFL, only a proportion is labeled. Fig. \ref{fig:exp1} illustrates the accuracy vary of the global FL model in each communication round. 
In addition, $\eta$ is set to 0.5 in this simulation, which means the data among the MUs is non-IID.

As shown in Fig. \ref{fig:exp1}, when $\gamma=0.1$, the accuracy of the XPFL is less than the CL and FedAvg algorithm, while is better than SemiFL with $\gamma=0.1$ or $\gamma=0.2$. 
When $\gamma=0.2$, the XPFL is only less than the CL algorithm and exceeds the other contenders including FedAvg. It is obvious that the accuracy of XPFL and SemiFL is increased with the increase of $\gamma$, while the accuracy improvement of XPFL is better than SemiFL, which could be seen from Fig. \ref{fig:exp1}(a) and (b).

Since the CL algorithm could use all labeled data for training without the model impairments from the transmitting parameters, its performance is always the best.
Although all the data used by FedAvg is also labeled, its performance is affected by the non-IID data. Thus, the XPFL with $\gamma=0.2$ has a competitive performance with the FedAvg on three datasets. 
We speculate the excellence of the proposed XPFL is mainly attributed to the GFed algorithm.
By combining the GAE-based unsupervised learning with KD-based semi-supervised learning, the unlabeled data is fully utilized. In addition, the personalized global aggregation relieves the impact of the non-IID data.

\subsection{Evaluation of XPFL for Non-IID}
\iffalse
\begin{figure}[htbp]
	\centering
	\includegraphics[width=9cm]{exp0.png}
	\caption{The data distribution on each MU under different $\eta$, where (a) $\eta=0.1$, (b) $\eta=0.3$, and (c) $\eta=0.5$.}
	\label{fig:exp0}
\end{figure}
\fi
This simulation evaluates the performance of the XPFL framework under the condition that the data in MUs is non-IID. 
In this subsection, we consider the following algorithms designed for the non-IID situation as the contenders:
\begin{itemize}
	\item FedAvgM \cite{hsu2019measuring}: An FL approach that combines the FedAvg and a mitigation strategy via server momentum.
	\item FedProx \cite{li2020federated}:
	A generalization and re-parametrization of FedAvg, which aims to tackle heterogeneity in federated networks.
	\item Scaffold \cite{karimireddy2020scaffold}:
	An FL approach adopts the use of control variates (variance reduction) to address the issue of ``client drift".
	\item XPFL: Our proposed novel FL approach, which solves the problem of non-IID data through personalized global aggregation.
\end{itemize}
\begin{figure*}[htbp]
	\centering
	\includegraphics[width=18cm]{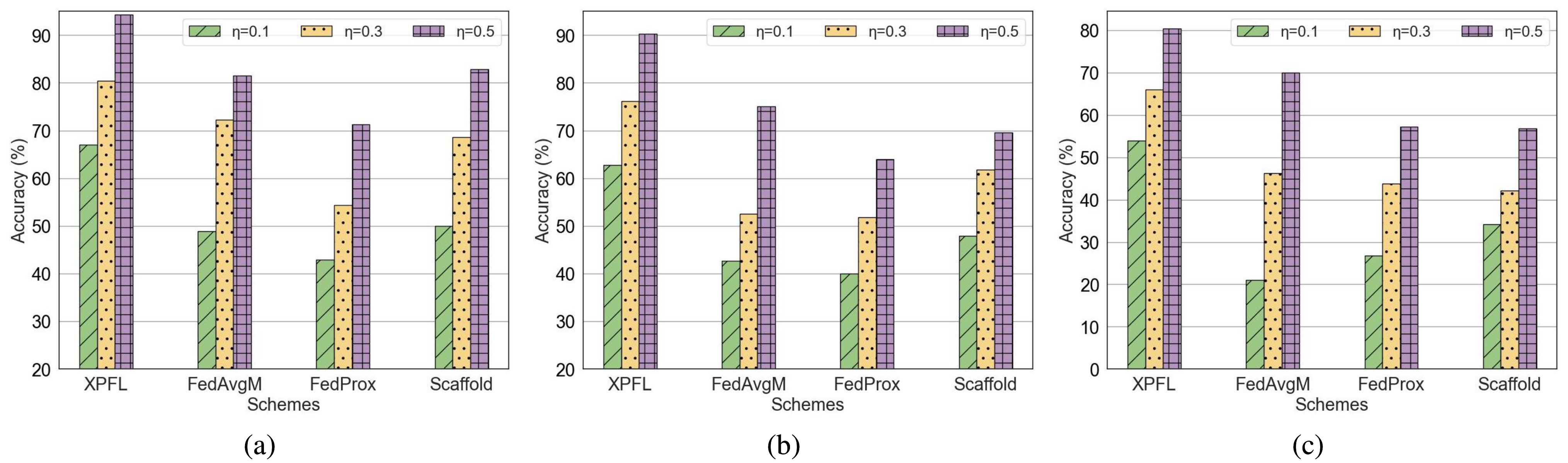}
	\caption{The final accuracy of the global FL model with different schemes on datasets (a) MNIST, (b) Fashion-MNIST, and (c) CIFAR-10.}
	\label{fig:exp2}
\end{figure*}

Note $\gamma$ is set to 0.1, which means all the above methods only have 10\% labeled data for local training in this simulation. 
In addition, we evaluate the performance of these schemes when $\eta = 0.1$, $\eta = 0.3$, and $\eta = 0.5$, respectively. 
%The data distribution of clients is displayed in Fig. \ref{fig:exp0}. We can see that the degree of non-IID increases with the decrease of $\eta$.

Fig. \ref{fig:exp2} shows the final accuracy of the global FL model obtained with the different FL schemes under different $\eta$. It is obvious that the accuracy of all schemes improves with the increase of $\eta$. We can see that on three datasets, the proposed XPFL method obtains the highest accuracy under different $\eta$. 
In particular, when $\eta=0.1$, the gap between the XPFL and the other contenders enlarges with the increase of the dataset's complexity. This could be seen obviously in Fig. \ref{fig:exp2}(a) and (c). 

For the superiority of the XPFL method, we suggest the application of the GFed algorithm is critical. 
Specifically, the GFed algorithm designs a cosine distance-based personalized global aggregation to allow local FL models to benefit from the aggregation and adapt to the respective local data set, thus overcoming the effect of non-IID data in FL. Moreover, the introduced GAI model can help the local FL model obtain extra knowledge from the unlabeled data, further improving the local FL model's performance.

\subsection{Evaluation of XPFL for Explainability}
\begin{figure*}[htbp]
	\centering
	\includegraphics[width=18cm]{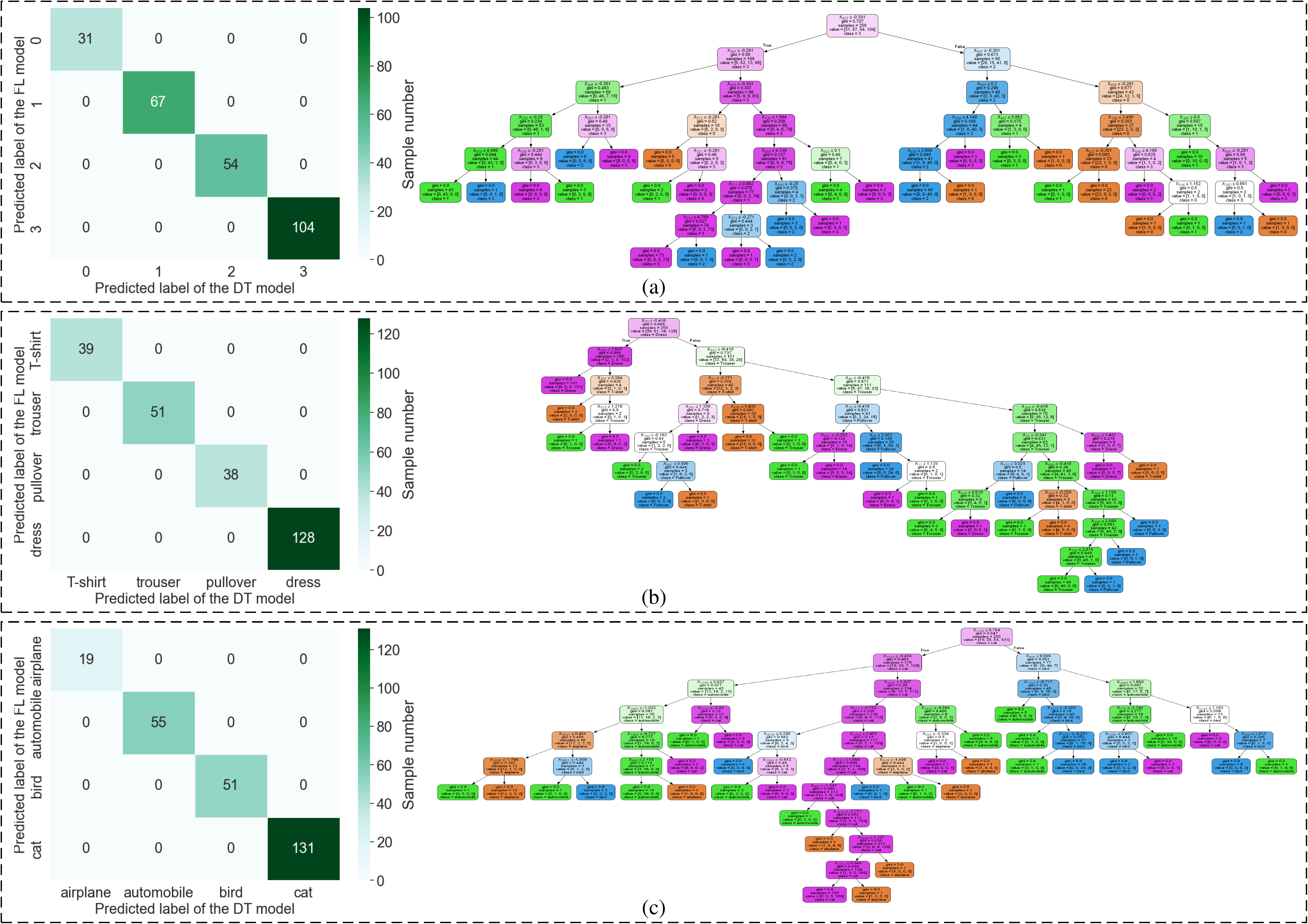}
	\caption{The illustration that the DT explains the FL model under different datasets (a) MNIST, (b) Fashion-MNIST, and (c) CIFAR-10. In each figure, on the left, the confusion matrix indicates the consistency of the prediction results of the DT and the FL model. On the right, the DT transparencies the decision process from input to output for each sample, where nodes of different colors represent different categories. The darkness of the color reflects the probability of samples belonging to this category in the node - the darker the color, the higher the probability.}
	\label{fig:exp3-1}
\end{figure*}
\begin{figure*}[htbp]
	\centering
	\includegraphics[width=18cm]{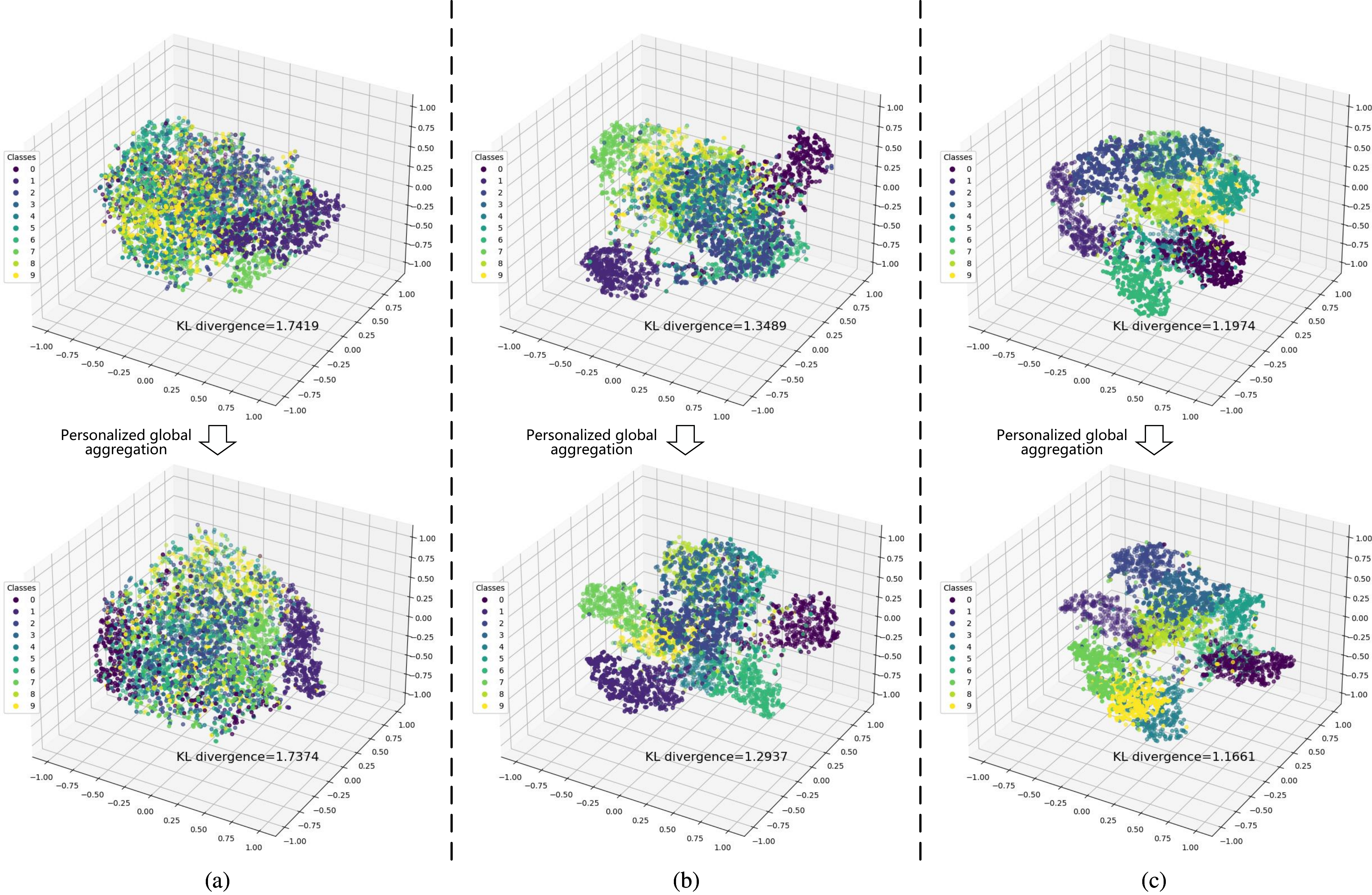}
	\caption{The illustration that the t-SNE explains the FL model before and after the personalized global aggregation in different communication rounds, where (a) $t=1$, (b) $t=20$, and (c) $t=40$. In each figure, the image above represents the explainable result of the local FL model before fusing it with the global FL model, while the image below is the explainable result after fusion.}
	\label{fig:exp3-2}
\end{figure*}

This simulation aims to display the explainability of XPFL in local training and global aggregation, respectively. %To facilitate the display of the experimental results, we only use less amount of data for the explainability simulation.   

Fig. \ref{fig:exp3-1} illustrates the result of the DT fitting the FL model and the explanation for predicting every input sample in someone's MU. Since the data is non-IID among MUs, each MU only has the data of part categories. Therefore, the test data of the three datasets only have four labels in this MU.
From the left heatmaps in Fig. \ref{fig:exp3-1}(a) - (c), we can see the DT perfectly fits the output of the FL model, which means the outputs of the DT and FL models are the same for the same input. Then, in the right of figure Fig. \ref{fig:exp3-1}(a) - (c), the decision process from all inputs to outputs is visualized through a DT.
In these DT figures, the leaf nodes with different colours represent different classification categories, and the number of categories is four.

Fig. \ref{fig:exp3-2} shows the explanation results of the FL model before and after the update through the t-SNE in different communication rounds. 
The upper and lower figures in Fig. \ref{fig:exp3-2}(a)-(c) are the visualized results before and after the update, respectively. 
Comparing the upper to the lower figures, whether from the value of the KL divergence or the distribution of analog data, we can see the fit of the FL model has an improvement after each model update.
In particular, from Fig. \ref{fig:exp3-2}(a), we can see the FL model fits the test data poorly when $t=1$, which means the quality of the FL model is bad. In Fig. \ref{fig:exp3-2}(b), the fit of the FL model has an improvement. In Fig. \ref{fig:exp3-2}(c), the FL model can distinguish the data sample from different categories well, which means the FL model can extract the data feature accurately. 

Fig. \ref{fig:exp3-1} and Fig. \ref{fig:exp3-2} demonstrate the explainability of the XPFL, which is mainly implemented by the XFed algorithm.
In the XFed algorithm, on the one hand, by regarding the DT as the explainer of the FL model, we can explain the relationship between any input and output of the FL model, thus achieving the transparency and explainability of the FL model's decisions. On the other hand, the t-SNE-based explainability method makes the changes of the local FL models during the global aggregation can be captured in a visualized way, thus ensuring the transparency and trustworthiness of the global aggregation.

\section{Conclusion}
To overcome the issues when FL is implemented in actual scenes, we propose the XPFL framework. 
First, we present the GFed algorithm to solve label scarcity and non-IID issues.
Specifically, in local training, we use the GAE to learn the massive unlabeled data, and then apply the KD-based semi-supervised learning to train the local FL with the assistance of GAE.
In global aggregation, after parameter aggregation, the new local model is obtained by fusing local and global models adaptively. 
This process aims to allow each local model to acquire knowledge from the others while maintaining its personalized characteristics.
Second, the XFed algorithm is designed to achieve FL explainability. 
In local training, we apply a DT to fit the input and output of the local FL model and use it as an explainer of the FL model.
In global aggregation, we employ the t-SNE to visualize each local FL model before and after updates. Thus, we can explain the change in each local FL model during the aggregation process.
Finally, the simulation demonstrates the validity of the proposed XPFL. 

In the future, we will consider the client selection strategy based on the explainability results, thus choosing the MUs with high value and saving the training cost. Additionally, we also study adding more complex communication and energy models into this framework, making it more reasonable and realistic.
%In the future, we will consider that design an explainable DL model, thus reducing the extra cost due to using the DT and t-SNE.
%\section{\textbf{References}}

\bibliographystyle{IEEEtran}
\bibliography{bare_jrnl_bobo}
\section*{Biographies}
\vspace{-10mm}
\begin{IEEEbiography}[{\includegraphics[width=1in,height=1.25in,keepaspectratio]{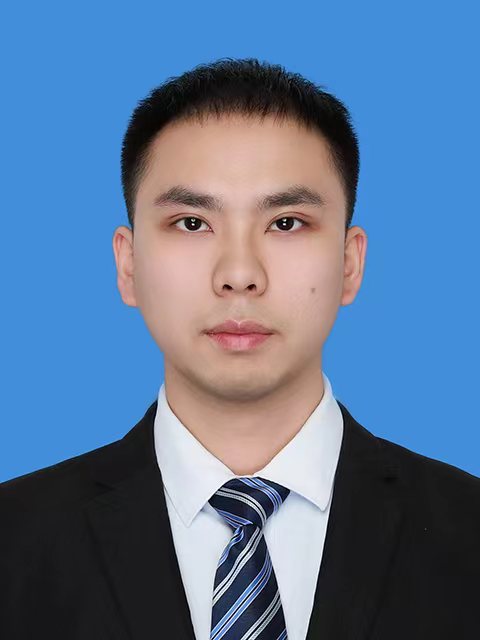}}]{Yubo Peng} received his B.S. and M.S. degrees from Hunan Normal University, Changsha, China, in 2019 and 2024. He is pursuing a Ph.D. from the School of Intelligent Software and Engineering at Nanjing University. His main research interests include semantic communication and generative artificial intelligence.
\end{IEEEbiography}

\vspace{-10mm}
\begin{IEEEbiography}[{\includegraphics[width=1in,height=1.25in,keepaspectratio]{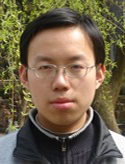}}]{Feibo Jiang} received his B.S. and M.S. degrees in the School of Physics and Electronics from Hunan Normal University, China, in 2004 and 2007, respectively. He received his Ph.D. degree in the School of Geosciences and Info-physics from the Central South University, China, in 2014. He is currently an associate professor at the Hunan Provincial Key Laboratory of Intelligent Computing and Language Information Processing, Hunan Normal University, China. His research interests include artificial intelligence, fuzzy computation, the Internet of Things, and mobile edge computing.
\end{IEEEbiography}

\vspace{-10mm}
\begin{IEEEbiography}[{\includegraphics[width=1in,height=1.25in,keepaspectratio]{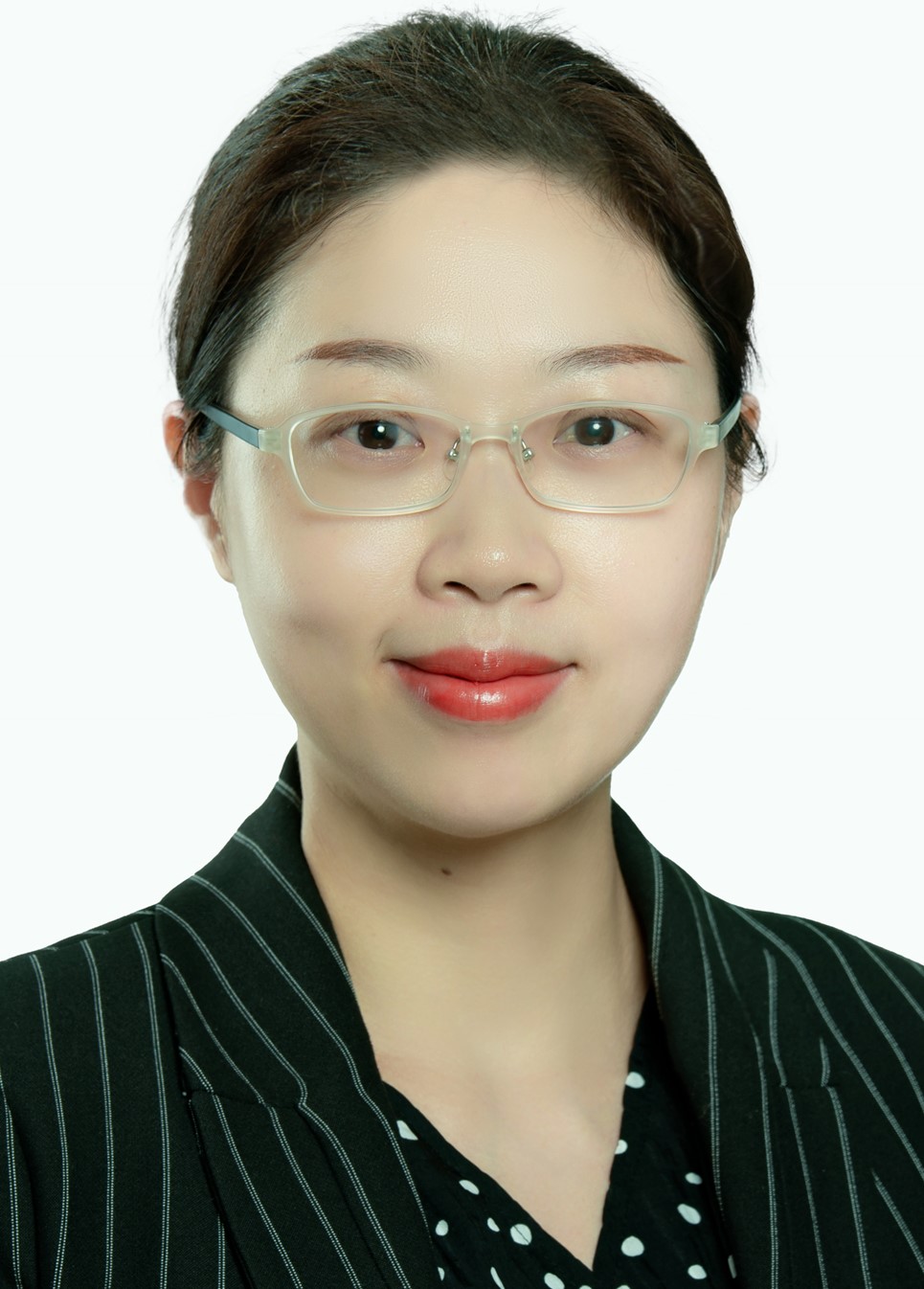}}]{Li Dong} received the B.S. and M.S. degrees in School of Physics and Electronics from Hunan Normal University, China, in 2004 and 2007, respectively. She received her Ph.D. degree in School of Geosciences and Info-physics from the Central South University, China, in 2018. Her research interests include machine learning, the Internet of Things, and mobile edge computing.
\end{IEEEbiography}

\vspace{-10mm}
\begin{IEEEbiography}[{\includegraphics[width=1in,height=1.25in,keepaspectratio]{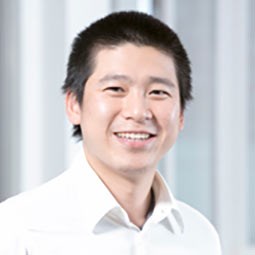}}]{Kezhi Wang}received the Ph.D. degree in Engineering from the University of Warwick, U.K. He was with the University of Essex and Northumbria University, U.K. Currently, he is a Senior Lecturer with the Department of Computer Science, Brunel University London, U.K. His research interests include wireless communications, mobile edge computing, and machine learning.
\end{IEEEbiography}

\vspace{-10mm}
\begin{IEEEbiography} [{\includegraphics[width=1in,height=1.25in,clip,keepaspectratio]{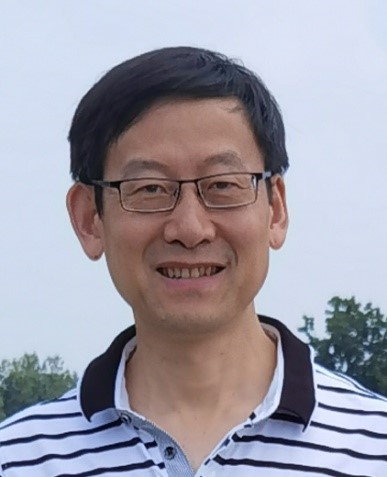}}]{Kun Yang} received his PhD from the Department of Electronic \& Electrical Engineering of University College London (UCL), UK. He is currently a Chair Professor in the School of Intelligent Software and Engineering, Nanjing University, China. He is also an affiliated professor of University of Essex, UK. His main research interests include wireless networks and communications, communication-computing cooperation, and new AI (artificial intelligence) for wireless. He has published 400+ papers and filed 30 patents. He serves on the editorial boards of a number of IEEE journals (e.g., IEEE WCM, TVT, TNB). He is a Deputy Editor-in-Chief of IET Smart Cities Journal. He has been a Judge of GSMA GLOMO Award at World Mobile Congress – Barcelona since 2019. He was a Distinguished Lecturer of IEEE ComSoc (2020-2021). He is a Member of Academia Europaea (MAE), a Fellow of IEEE, a Fellow of IET and a Distinguished Member of ACM.
\end{IEEEbiography} 
\newpage
\end{document}

\title{Supplementary Material}

\markboth{Submitted for Review}%
{Shell \MakeLowercase{\textit{et al.}}: Bare Demo of IEEEtran.cls for IEEE Journals}



\maketitle 


% \fi
\IEEEpeerreviewmaketitle

\section{Supplementary Convergence and Computational Complexity Analysis}
\subsection{Convergence Analysis}
This subsection presents the convergence analysis of XPFL, with fixed $\psi_{k,t}$, on a strongly convex function.
For the better subsequent analysis, we give the following definitions \cite{woodworth2020minibatch}:
\newtheorem{definition}{Definition}[]
\begin{definition}\label{df1}
	(Gradient diversity). Diversity among the global gradients, in relation to the local gradient of the device $k$, is measured using the following quantity:
	\begin{equation}
		\zeta_{k}=\sup_{w_{g,t}}\rVert\nabla F_{g,t}\left( w_{g,t} \right)-F_{k,t}(w_{g,t})\rVert^2,
	\end{equation}
	where $\nabla F_{g,t}(\cdot)$ denotes the global gradient in the $t$-th communication round, while $\nabla F_{k,t}(\cdot)$ represents the local gradient of device $k$. 
\end{definition}

\begin{definition}\label{df2}
	(Local-global optimality gap) The discrepancy between the optimal local and global FL models is quantified using the following quantity:
	\begin{equation}
		\Delta_k=\rVert w_k^*-w_g^*\rVert^2,
	\end{equation}
	where $w_k^*$ denotes the optimal local FL model in the device $k$, whereas $w_g^*$ is the optimal global FL model.
\end{definition}
Definition \ref{df1} and \ref{df2} reflect the data heterogeneity between different devices.

Then, the standard assumptions regarding the analytical properties of loss functions are defined to obtain convergence theory, as discussed in \cite{woodworth2020minibatch}:

\newtheorem{assumption}{Assumption}[]
\begin{assumption}\label{ass1}
	(Smoothness). There exists a $L > 0$ such that:
	\begin{equation}
		\rVert\nabla F_{k,t}\left( w_{k,t1} \right)-\nabla F_{k,t}(w_{k,t2})\rVert\leq L\rVert w_{k,t1}-w_{k,t2}\rVert, \forall k \in \mathcal{K},
	\end{equation}
	where $w_{k,t1}$ and $w_{k,t2}$ represent the local FL models of the device $k$ in the communication round $t1$ and $t2$, respectively.
\end{assumption}

\begin{assumption}\label{ass2}
	(Bounded variance). The variance of stochastic gradients computed at each local data shard is bounded:
	\begin{equation}
		\mathbb{E}[\rVert\nabla F_{k,t}\left( w_{k,t1};\xi \right)-\nabla F_{k,t}(w_{k,t2})\rVert^2]\leq \varrho^2, \forall k \in \mathcal{K},
	\end{equation}
	where $F_{k,t}\left( \cdot;\xi \right)$ denotes the stochastic gradient of $F_{k,t}\left( \cdot \right)$ computed at mini-batch $\xi$.
\end{assumption}

\begin{assumption}\label{ass3}
	(Strong convexity). There exists a $\mu > 0$ satisfies:
	\begin{equation}
		\begin{split}
			&F_{k,t}(w_{k,t1}) \geq F_{k,t}(w_{k,t2}) + \langle\nabla F_{k,t}(w_{k,t2}), w_{k,t2}-w_{k,t1}\rangle \\
			&+ \frac{\mu}{2}\rVert w_{k,t1}-w_{k,t2}\rVert^2, \forall k \in \mathcal{K},
		\end{split}
	\end{equation}
	where $\langle \cdot \rangle$ represents the inner product operation. 
\end{assumption}

The XPFL framework updates the local FL model by incorporating the partial global FL model, with the goal of finding the convergence rate of the updated local FL model. Specifically, we denote $\hat{w}_{k,t}$ as the personalized local FL model, which is obtained by scaling the local model $w_{k,t}$ with a weighting factor $\psi_{k,t}$ and adding to it the complement of the factor times the global model $w_{g,t}$. The convergence analysis determines the impact of the global FL model on the personalized local FL model in terms of gradient diversity, personalized model convergence, global model convergence, and local-global optimality gap. The distance between gradients of the personalized local FL model and the global FL model can be formulated as: 
\begin{equation}
	\begin{split}
		&\mathbb{E}[\rVert\nabla F_{k,t}\left( \hat{w}_{k,t}\right)-\nabla F_{g,t}(w_{g,t})\rVert^2]\leq 6\zeta_{k} + \\
		&2L^2\mathbb{E}[\rVert\hat{w}_{k,t}-\hat{w}_k^*\rVert^2] + 6L^2\mathbb{E}[\rVert w_{g,t}-w_g^*\rVert^2] + 6L^2\Delta_k,
	\end{split}
\end{equation}
where $\hat{w}_k^*$ is the optimal personalized FL model. 
$\mathbb{E}[\rVert\hat{w}_{k,t}-\hat{w}_k^*\rVert^2]$ and $\mathbb{E}[\rVert w_{g,t}-w_g^*\rVert^2]$ can converge rapidly under smooth strongly convex objectives. Additionally, $\zeta_{k}$ and $\Delta_k$ are utilized to account for heterogeneity among local loss functions.
The preceding analysis leads to the following theorems:

\newtheorem{theorem}{Theorem}[]
\begin{theorem}\label{theo1}
	(Global model convergence of XPFL). When each device’s loss function satisfies Assumptions \ref{ass1}, \ref{ass2}, and \ref{ass3}, by setting the learning rate $\varepsilon=\frac{16}{\mu(t+a)}$, where $a=\max\{128\kappa,G\}$ and $\kappa=\frac{L}{\mu}$, the convergence of the global FL model can be expressed as \cite{deng2020adaptive}:
	\begin{equation}
		\begin{split}
			&\mathbb{E}[F_{g,t}(w_{g,t})]-F_g^* \leq \mathcal{O}(\frac{\mu\mathbb{E}[\rVert w_{g,0}-w_g^*\rVert^2]}{T^3}) + \\
			&\mathcal{O}(\frac{\kappa^2 G(\varrho^2+2G\frac{\zeta}{K})}{\mu T^2}) +
			\mathcal{O}(\frac{\varrho^2}{KT}),
		\end{split}
	\end{equation}
	where $F_g^*$ represents the minimum of $F_{g,t}$; $w_{g,0}$ refers to the untrained global FL model; $\zeta=\sum_{k=1}^{K}\zeta_k$; $\mathcal{O}(\cdot)$ is used to further hide constants.
\end{theorem}

\begin{theorem}\label{theo2}
	Suppose each device’s local loss function satisfies Assumptions \ref{ass1}, \ref{ass2}, and \ref{ass3}, and set as $\kappa=\frac{L}{\mu}$. We assume $\psi_{k,t}\geq \max\{1-\frac{1}{4\sqrt{6}\kappa},1-\frac{1}{4\sqrt{6}\kappa\sqrt{\mu}}\}$. By setting learning rate $\varepsilon=\frac{16}{\mu(t+a)}$, where $a=\max\{128\kappa,G\}$ and $\kappa=\frac{L}{\mu}$, the convergence of the personalized local FL model can be given by \cite{deng2020adaptive}: 
	\begin{equation}
		\begin{split}
			&\mathbb{E}\left[F_{k,t}\left(\hat{w}_{k,t}\right)\right]-F_k^{*} =\mathcal{O}\left(\frac{\mu}{ T^{3}}\right)+\psi_{k,t}^{2} \mathcal{O}\left(\frac{\varrho^{2}}{\mu T}\right)+\\
			&\left(1-\psi_{k,t}\right)^{2}\left\{ \mathcal{O}\left(\frac{\zeta_{k}}{\mu}+\frac{\kappa L \Delta_{k}}{2}\right)+\mathcal{O}\left(\frac{\kappa L \ln T}{ T^{3}}\right)+\right.\\
			&\left.\mathcal{O}\left(\frac{\kappa^{2} \varrho^{2}}{\mu K T}\right)\right\}+\left(1-\psi_{k,t}\right)^{2}
			\mathcal{O}\left(\frac{\kappa^{2} G\left(\varrho^{2}+G\left(\zeta_{k}+\frac{\zeta}{K}\right)\right)}{\mu T^{2}}+\right.\\
			&\left.\frac{\kappa^{4} G\left(\varrho^{2}+2 G \frac{\zeta}{K}\right)}{\mu T^{2}}\right)
		\end{split}
	\end{equation}
	where $F_k^*$ represents the minimum of $F_{k,t}$.
\end{theorem}

Based on the above theorem, we can get the following corollary:
\newtheorem{corollary}{Corollary}[]
\begin{corollary}
	In Theorem \ref{theo2}, when we set as $G=\sqrt{T/K}$, we can obtain the convergence rate:
	\begin{equation}
		\begin{split}
			&\mathbb{E}\left[F_{k,t}\left(\hat{w}_{k,t}\right)\right]-F_k^{*} \leq \psi_{k,t}^{2} \mathcal{O}\left(\frac{\varrho^{2}}{\mu T}\right)+\\
			&\left(1-\psi_{k,t}\right)^{2}\left(\mathcal{O}\left(\frac{\kappa^{2} \varrho^{2}}{\mu K T}\right)+\mathcal{O}\left(\frac{\kappa^{2}\left(\zeta_{k}+\frac{\zeta}{K}\right)}{\mu K T}\right)\right)+\\
			&\left(1-\psi_{k,t}\right)^{2} \mathcal{O}\left(\frac{\zeta_{k}+\frac{\zeta}{K}}{\mu}+\kappa L \Delta_{k}\right).
		\end{split}
	\end{equation}
\end{corollary}

\subsection{Computational Complexity Analysis}
In local training, the extra computation, explaining the local FL model, is determined by the DT's construction cost. 
The computational complexity of DT construction is determined by the criterion function, which involves two primary operations: calculating entropy gain for the class attribute and computing entropy for each input sample with respect to the class \cite{sani2018computational}.
The cost of computing probability for each class depends on the sample size and is therefore $O(n)$, where 
$n$ is the size of the training dataset. The cost of computation on one input attribute is $O(n\log_{2}n)$. Considering all attributes leads to a total cost of $O(mn\log_{2}n)$, where
$m$ is the number of attributes.
Since instances and their target values are considered at each split, the estimated complexity for this operation is $O(n\log_{2}n)$ when recursive calls are made on subsets of the training set.
Therefore, the computational complexity of building a DT is $O(n)+O(mn\log_{2}n)+O(n\log_{2}n)$.

In global aggregation, the extra computation, explaining the FL model, is determined by the t-SNE method. And the t-SNE has a computational complexity of $O(n^2)$ \cite{van2008visualizing}.

Based on the above analysis, the computational complexity for achieving the FL explainability is mainly bounded by the training data set. The training data set size is larger, the explainability is stronger, and the extra computation cost is higher. Hence, we should consider how to trade off the relationship between the explainability and computation cost in the actual scene.

\section{Supplementary Data Distribution on Different Clients}
%\section{\textbf{References}}
\begin{figure}[htbp]
	\centering
	\includegraphics[width=9cm]{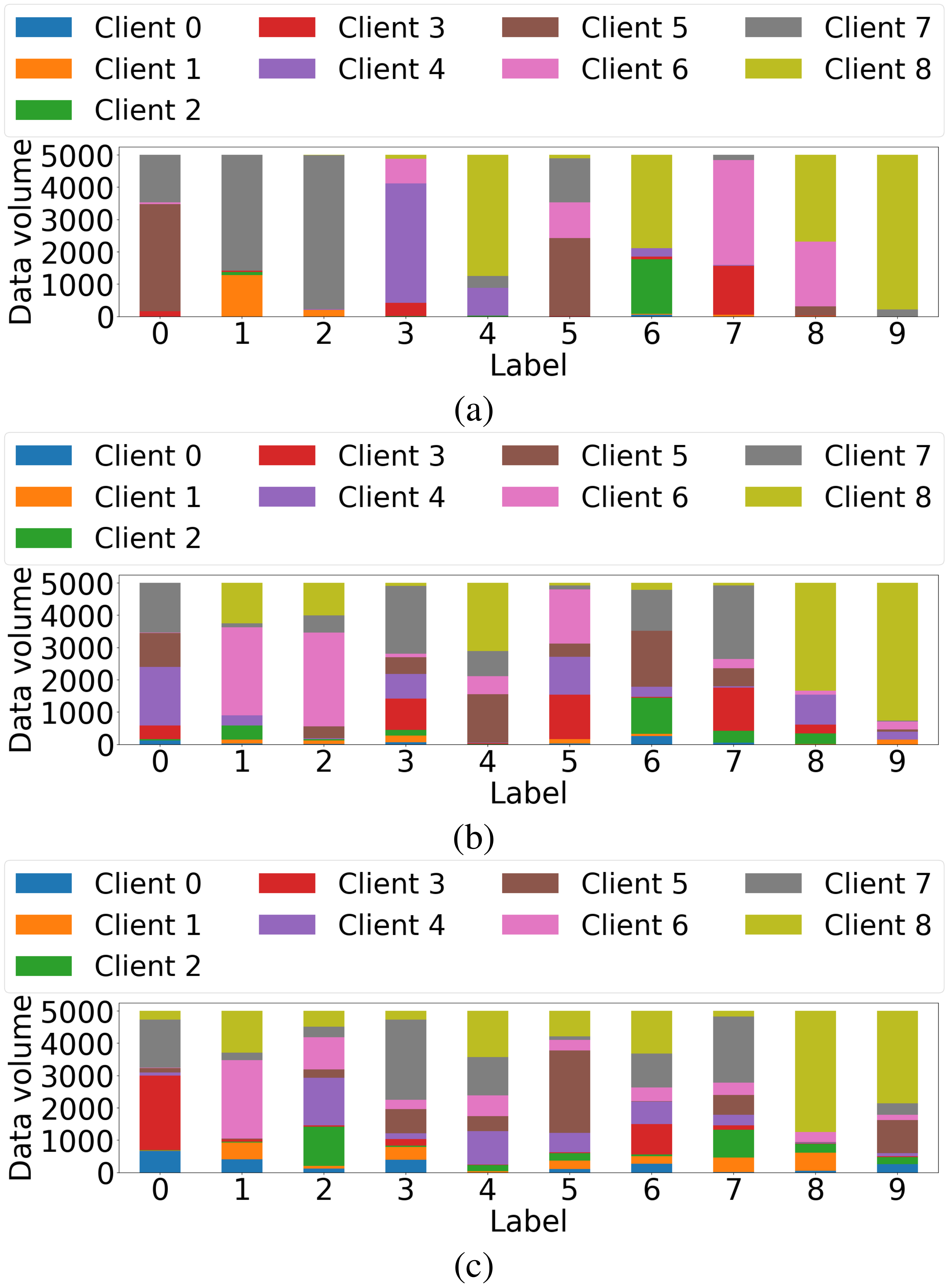}
	\caption{The data distribution on each client under different $\eta$, where (a) $\eta=0.1$, (b) $\eta=0.3$, and (c) $\eta=0.5$.}
	\label{fig:exp0}
\end{figure}

Fig. \ref{fig:exp0} illustrates the data distribution of clients on CIFAR-10. 
It is evident that the degree of non-IID (non-identically and independently distributed data) varies with $\eta$. A smaller value of $\eta$ corresponds to a greater disparity among clients, whereas a larger $\eta$ results in a more similar data distribution among clients.

\bibliographystyle{ieeetran}
\bibliography{bare_jrnl_bobo}
\newpage